\documentclass[10pt,journal,compsoc]{IEEEtran}
\ifCLASSOPTIONcompsoc
  \usepackage[nocompress]{cite}
\else
  \usepackage{cite}
\fi
\ifCLASSINFOpdf
\else
\fi
\usepackage[utf8]{inputenc} 
\usepackage{hyperref}       
\usepackage{url}            
\usepackage{booktabs}       

\usepackage{floatrow}
\newfloatcommand{capbtabbox}{table}[][0.93\FBwidth]
\usepackage{marvosym}
\usepackage{graphicx}
\usepackage[lined,boxed,commentsnumbered,ruled]{algorithm2e}
\usepackage{multirow}
\usepackage{colortbl}
\usepackage{float}
\usepackage{subfigure}
\usepackage{makecell}
\usepackage{caption}

\newcommand{\jn}[1]{{#1}}
\usepackage{amsmath, amsthm, amssymb, multirow, paralist}
\newtheorem{thm}{Theorem}

\newtheorem{assumption}{Assumption}

\def \U {\mathcal{U}}
\def \E {\mathrm{E}}
\def \x {\mathbf{x}}

\def \xl {\mathbf{x}^{l}}
\def \xu {\mathbf{x}^{u}}

\def \Lmath {\mathcal{L}} %

\def \q {\mathbf{q}}

\def \p {\mathbf{p}}
\def \q {\mathbf{q}}

\def \E {\mathrm{E}}%

\def \x {\mathbf{x}}
\def \y {\mathbf{y}}

\def \Lt {\widehat{\mathcal L}} %
\def \Dt {\widehat{\mathcal D}} %
\def \gt {\widehat{g}}%
\def \gi {g_{\text{id}}}
\def \gf {g_{\text{uf}}} 
\def \gn {g_{\text{fr}}} 

\def \gxu {g_{\mathbf{x}^{u}_{i}}} 

\usepackage[normalem]{ulem}

\begin{document}
%
\title{Robust Semi-supervised Learning by\\ Wisely Leveraging Open-set Data}
%
%
%
%

\author{Yang~Yang,~\IEEEmembership{Member,~IEEE,}
        Nan~Jiang,
        Yi~Xu,
        and~De-Chuan~Zhan
\IEEEcompsocitemizethanks{\IEEEcompsocthanksitem Yang~Yang is with the school of Computer Science and Engineering, Nanjing University of Science and Technology, 	Nanjing 210094, China. \protect\\
E-mail: yyang@njust.edu.cn
\IEEEcompsocthanksitem Nan~Jiang and De-Chuan~Zhan are with the National Key Laboratory for Novel Software Technology, Nanjing University, and also with the School of Artificial Intelligence, Nanjing University, Nanjing 210023, China. \protect\\
E-mail: jiangn@lamda.nju.edu.cn, zhandc@nju.edu.cn
\IEEEcompsocthanksitem Yi~Xu is with the School of Control Science and Engineering, Dalian University of Technology, Dalian 116081, China. \protect\\
E-mail: yxu@dlut.edu.cn}
\thanks{Corresponding authors: Yi~Xu and De-Chuan~Zhan.}
}

\markboth{}
{Shell \MakeLowercase{\textit{et al.}}: Bare Demo of IEEEtran.cls for Computer Society Journals}
%



\IEEEtitleabstractindextext{%
\begin{abstract}
Open-set Semi-supervised Learning (OSSL) holds a realistic setting that unlabeled data may come from classes unseen in the labeled set, \emph{i.e.}, out-of-distribution (OOD) data, which could cause performance degradation in conventional SSL models. To handle this issue, except for the traditional in-distribution (ID) classifier, some existing OSSL approaches employ an extra OOD detection module to avoid the potential negative impact of the OOD data. Nevertheless, these approaches typically employ the entire set of open-set data during their training process, which may contain data unfriendly to the OSSL task that can negatively influence the model performance. This inspires us to develop a robust open-set data selection strategy for OSSL. Through a theoretical understanding from the perspective of learning theory, we propose \textbf{Wise} \textbf{Open}-set Semi-supervised Learning (WiseOpen), a generic OSSL framework that selectively leverages the open-set data for training the model. By applying a gradient-variance-based selection mechanism, WiseOpen exploits a friendly subset instead of the whole open-set dataset to enhance the model's capability of ID classification. Moreover, to reduce the computational expense, we also propose two practical variants of WiseOpen by adopting low-frequency update and loss-based selection respectively. Extensive experiments demonstrate the effectiveness of WiseOpen in comparison with the state-of-the-art.
\end{abstract}

}

\maketitle

\IEEEdisplaynontitleabstractindextext

\IEEEpeerreviewmaketitle

\IEEEraisesectionheading{\section{Introduction}\label{intro}}

\IEEEPARstart{S}{emi-supervised} learning (SSL)~\cite{CSZ2006, zhu2009introduction} leverages the ubiquitous unlabeled data to break the limitation of supervised learning (SL) caused by the huge human and financial costs in obtaining labeled data~\cite{li2019towards, zhou2018brief, yang2021semi1, yang2023cost}. There exist various techniques for SSL, such as consistency regularization~\cite{LaineA17, MiyatoMKI19, TarvainenV17} and entropy minimization~\cite{GrandvaletB04, lee2013pseudo}. Moreover, some recent holistic approaches~\cite{BerthelotCGPOR19, remixmatch, SohnBCZZRCKL20, XuSYQLSLJ21, yang2021s2osc, zhao2022lassl} which integrate the techniques from dominant SSL paradigms have successfully achieved excellent performance on many benchmarks. 

\begin{figure}[!tb]
    \centering
    \includegraphics[width=0.90\textwidth]{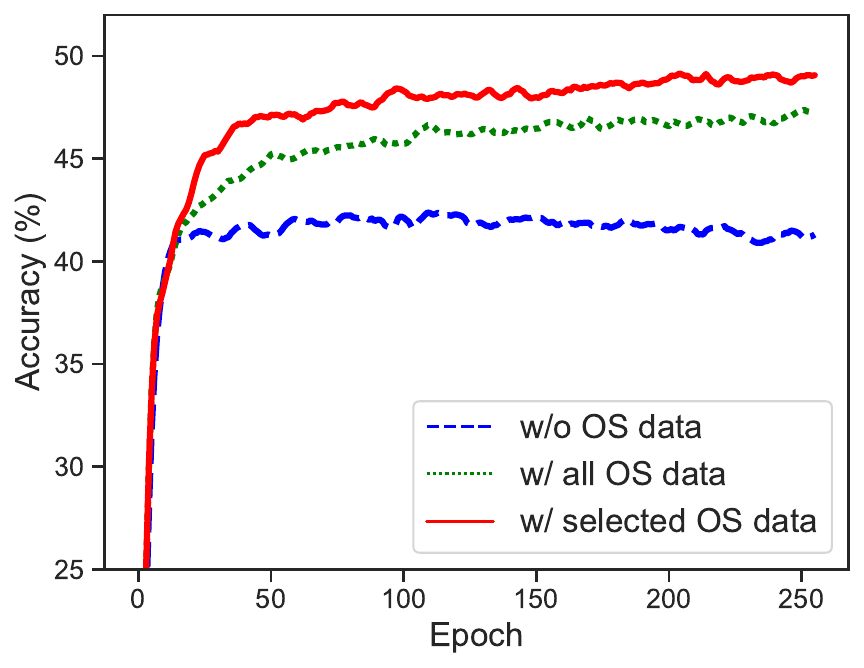}
    \caption{An example of models' performance (testing accuracy on ID classification) with different strategies of using the \textbf{o}pen-\textbf{s}et data (OS data) illustrates the effectiveness of selectively leveraging OS data during the training process. Experiments are conducted on Tiny-ImageNet at 120 seen classes with 50 labels for each class. We employ the following methods: (1) Labeled Only (w/o OS data), an SL method only trained with labeled data; (2) OpenMatch~\cite{saito2021openmatch} (w/ all OS data), an OSSL method trained with all OS data; and (3) WiseOpen-L on top of OpenMatch(w/ selected OS data), an OSSL method trained with selected OS data.}
    \label{fig:motivation}
\end{figure}

Despite all these achievements acquired by SSL, traditional SSL typically makes the assumption that labeled data and unlabeled data share the same class space~\cite{yang2024exploiting,yang2021semicluster}. However, in real applications, the unlabeled training dataset may contain the data from classes unseen in the labeled, \emph{i.e.}, OOD data, which may induce the existing SSL models to overconfidently misclassify data from unseen classes to nearby seen classes~\cite{guorobust, xi2023robust, yang2023not}. Aiming at promoting SSL to more realistic scenarios, OSSL~\cite{ChenZLG20, guo2020safe, saito2021openmatch, yu2020multi} has been widely investigated. An ideal OSSL model should have the capability of tackling the following task: classifying the ID testing instances under the potential interference of the open-set training data. Most existing OSSL approaches~\cite{guo2020safe, saito2021openmatch, yu2020multi} usually apply an OOD detection module in addition to the traditional ID classifier, for the purpose of acquiring the capability of differentiating OOD data from ID data, thereby avoiding their potential negative impacts on ID classifier training. Typically, all open-set (OS) data are involved in these OSSL models' training, which may comprise both friendly data and unfriendly data. Here \emph{friendly data} means the data are beneficial to the considered OSSL task, while unfriendly can be the opposite. Due to the possible existence of unfriendly data, the model effectiveness may be affected if recklessly use the complete OS data set. On the other hand, discarding all OS training data is also not a wise choice, since it can result in the loss of valuable information contained in friendly OS data, ultimately leading to unsatisfactory performance. We can observe from Figure~\ref{fig:motivation} that the method using all OS data has better ID classification accuracy compared with the method without using OS data, which reflects that in this example, certain OS data (friendly data) can enhance the ID classification accuracy. Furthermore, the method with selected OS data has the best performance on ID classification. That is to say, in this example, excluding certain unfriendly OS data can further improve the ID classification performance, revealing that the selection of OS training data is essential for the OSSL task. Moreover, to give an insight into what data can be friendly and to better demonstrate that utilizing friendly OS data can enhance model effectiveness while using unfriendly OS data can lead to performance degradation, we derive a theoretical analysis in Section~\ref{sec:thm:und}. Inspired by both these empirical findings and theoretical analyses, we aim to carefully leverage selected OS data in OSSL. 

To this end, we propose a generic OSSL framework called \textbf{Wise} \textbf{Open}-set Semi-supervised Learning (\textbf{WiseOpen}), which selectively exploits the OS data depending on the gradient variance. Specifically, based on the theoretical analysis in Section~\ref{sec:thm:und} that friendly open-set data can promote better generalization, for every epoch, WiseOpen first calculates the loss gradient of the limited labeled data as the approximate expectation of the gradient of the loss. Then based on the approximate gradient expectation and the gradient of each open-set instance, it computes the gradient variance for each open-set instance, and open-set data with smaller gradient variance will be selected as friendly data to train the OSSL model along with the limited labeled data. By applying this Gradient-Variance-based Selection Mechanism (\textbf{GV-SM}), the model can achieve better performance by the wiser exploitation of the open-set data. Nevertheless, it is of high time cost to calculate the gradient variance for each instance in every epoch. To handle this tricky situation, we provide two practical and economic variants of WiseOpen named \textbf{WiseOpen}-\textbf{E}conomic and \textbf{WiseOpen}-\textbf{L}oss. WiseOpen-E sets a larger interval of updating the friendly open-set data set obtained by GV-SM, which can effectively make a balance between the time cost and the model performance. But meanwhile, WiseOpen-E will inevitably suffer from the stale selection issue which may harm the model performance. On the other hand, WiseOpen-L employs loss values as a substitute for the gradient variance to select friendly data, which can address the time-consuming problem without raising the stale-selecting problems. WiseOpen-L can lead to inclusiveness of some previous SSL and OSSL methods using certain loss-based or confidence-based selection mechanisms, like~\cite{SohnBCZZRCKL20, Wang0HHFW0SSRS023, XuSYQLSLJ21}. We theoretically demonstrate the rationality and feasibility of replacing GV-SM in WiseOpen and WiseOpen-E with Loss-based Selection Mechanism (\textbf{L-SM}) in WiseOpen-L. Experiments on CIFAR-10/100~\cite{krizhevsky2009learning} and Tiny-ImageNet~\cite{yao2015tiny} show that our naive approach, \emph{i.e.}, WiseOpen can achieve outstanding performance on the OSSL tasks by wisely leveraging the OS data while the two variants can also outperform the baselines. To summarize, our main contributions are:
\begin{itemize}
    \item  From the perspective of learning theory, we put forward an insight into the necessity of selectively leveraging the friendly open-set data in OSSL scenarios.
    \item We propose a robust general OSSL framework WiseOpen that employs GV-SM to wisely select friendly open-set data. This provides the OSSL community with a plug-and-play module to enhance the models' performance.
    \item We further provide WiseOpen-E and WiseOpen-L as two practical variants of WiseOpen, which can make the selection procedure more computation-friendly while still yielding performance improvements.
    \item The effectiveness of our proposed WiseOpen and its variants is demonstrated by extensive experiments on three popular benchmark datasets. 
\end{itemize}
\section{Related Work}
\label{rw}

\textbf{Semi-supervised Learning.} SSL aims at leveraging the unlabeled data to improve the model's performance without the extra cost of data annotation. With the advancement of deep learning~\cite{LeCunBH15}, a number of deep SSL methods have been reported. For example, entropy minimization methods~\cite{lee2013pseudo} focus on preventing the model from producing a flat prediction. Consistency regularization methods~\cite{LaineA17, MiyatoMKI19, TarvainenV17, XieDHL020} encourage the model to output the same results between differently augmented inputs. Additionally, holistic approaches like MixMatch~\cite{BerthelotCGPOR19} and FixMatch~\cite{SohnBCZZRCKL20} successfully integrate some prior SSL techniques in a framework to gain better performance. Specifically, FixMatch is a simple but effective method that jointly employs consistency regularization and pseudo-labeling techniques and applies a fixed threshold for selecting high-confident unlabeled data to train the model. After the success of FixMatch, recent works like Dash~\cite{XuSYQLSLJ21}, FlexMatch~\cite{ZhangWHWWOS21}, and FreeMatch~\cite{Wang0HHFW0SSRS023} further explore how to determine the suitable confidence thresholds according to model’s learning status so that better exploit unlabeled data for better performance. Despite the achievements acquired by these SSL methods, they typically hold the assumption that the labeled data and unlabeled data share the same class space. When it comes to the realistic open-set scenario in this paper, they usually fail to do the job, as they may misclassify the data from unseen classes to certain nearby seen classes overconfidently.

\noindent\textbf{Open-set Semi-supervised Learning.} OSSL, an emerging branch of SSL, considers a more realistic scenario that the unlabeled data may come from unseen classes except for seen classes. A variety of methods~\cite{realmix2019, yu2020multi,ChenZLG20, guo2020safe, HuangF0CWWLL21, saito2021openmatch, ParkYJS22, huang2022they, li2023iomatch} has been proposed in recent years. For example, D3SL~\cite{guo2020safe} applies meta-learning to obtain a weight function for alleviating the impact of OOD data. MTC~\cite{yu2020multi} employs a multi-task curriculum framework to leverage the unlabeled data in calculating the MixMatch~\cite{BerthelotCGPOR19} loss. OpenMatch~\cite{saito2021openmatch} adopts one-vs-all classifiers with soft open-set consistency regularization and incorporates FixMatch to handle the OSSL tasks. OpenCos~\cite{ParkYJS22} utilizes the model pre-trained by contrastive learning to identify the pseudo-ID and pseudo-OOD data which will be used for fine-tuning the pre-trained model with certain SSL loss plus an auxiliary loss that assigns soft labels to pseudo-OOD data. Although these methods have been reported to get respectable performance, they typically utilize all open-set data to train the models, which can lead to performance degradation caused by unfriendly open-set data. \jn{One recent method IOMatch~\cite{li2023iomatch} has proposed to select confident pseudo-ID data while calculating unlabeled inlier loss, and exclude unlabeled data with low confidence in open-set predictions while calculating open-set loss where all unseen classes are regarded as one single class. In other words, IOMatch separately designs the selection mechanism specialized for each single loss based on the prediction confidence, aiming at conquering the dilemma of unreliable results in the training procedure. In contrast, with an insight into selecting friendly data from the whole learning task perspective, we propose applying GV-SM or L-SM over the unified unsupervised loss, which is more generic, and is orthogonal and complementary to prior methods.}

\noindent\textbf{Out-of-distribution Detection.} OOD detection aims at detecting outliers that belong to the different distributions from the training distributions while ensuring that the ID classification is not adversely affected~\cite{yangopenood}. Existing OOD detection methods~\cite{DuWCL22, hendrycksbaseline, Shreyas2020ova, sun2022out, WeiXCF0L22, yang2023learning} have been reported to achieve excellent performance in tackling this task. However, these methods usually acquire abundant labeled data from seen classes and some methods~\cite{HendrycksMD19, YangWFYZZ021, YuA19} even additionally utilize external OOD knowledge in the training phase. In contrast, OSSL only has limited access to the labeled data of seen classes and needs to handle the unlabeled data from unseen classes in training, which renders the task more challenging.

\section{WiseOpen: Wisely Leverage the Open-set Data}
\label{proposal}
In this section, we first provide the preliminary of our work. Then through a theoretical understanding, we demonstrate the effectiveness of excluding unfriendly open-set data and utilizing friendly open-set data. Based on this, we then specify the selection mechanisms applied in our proposed approaches, which can help models gain better performance in ID classification by wisely leveraging the open-set data. Moreover, at the end of this section, we will clarify the relationships between our proposed frameworks and the existing OSSL methods.

\begin{table}[!htbp]
    \centering
    \small
    \begin{tabular}{lp{0.62\textwidth}}
        \toprule
        Notation &  \makecell[c]{Mathematical Meaning}\\
        \midrule
        $K$ & Number of the seen classes.\\
        $\mathcal{S}=\left\{(\xl_{i},\y_{i})\right\}_{i=1}^{N_{l}}$ & Labeled training dataset containing $N_{l}$ labeled pairs$(\xl_{i},\y_{i})$. \\
        $\U=\left\{\xu_{i}\right\}_{i=1}^{N_{u}}$ & Original unlabeled training dataset containing $N_{u}$ instance $\xu_{i}$. \\
        $\U_t$ & Selected unlabeled subset in the $t$-th epoch. \\
        $\mathcal{L}, \mathcal{L}_{s},\mathcal{L}_{u}$ & The overall loss, supervised loss, and unsupervised loss.\\
        $\theta$ & Parameters of the model. \\
        $g(\theta)$ & Stochastic gradient of loss function computed at $\theta$. \\
        $\E[\cdot]$ & Mathematical expectation of some random variable.\\
        $\text{ERB}(\cdot)$ & The excess risk bound given the model parameters.\\
        $|\cdot|$ &  Cardinality of the given set.\\
        \bottomrule
    \end{tabular}
    \caption{Frequently used notations along with their mathematical meaning.}
    \label{tb:notation}
\end{table}

\subsection{Preliminary}
\label{sec:pre}

Let $\mathcal{S}=\left\{(\xl_{i},\y_{i})\right\}_{i=1}^{N_{l}}$ be the labeled training data set, 
where $\xl_{i}$ is an ID example from one of the $K$ seen classes and $\y_{i}$ is its one-hot label. Let $\U=\left\{\xu_{i}\right\}_{i=1}^{N_{u}}$ be the unlabeled training data set, where $\xu_{i}$ is an unlabeled training instance drawn from the seen classes or unseen classes. Typically, the overall objective function $\mathcal{L}$ of OSSL, no matter what specific techniques are applied, can be written as 
\begin{equation}
    \label{eqn:loss:overall}
    \mathcal{L} = \mathcal{L}_{s}(\theta;\mathcal{S}) + \mathcal{L}_{u}(\theta;\U),
\end{equation}
where $\mathcal{L}_{s}$ and $\mathcal{L}_{u}$ represent supervised loss and unsupervised loss respectively and $\theta$ is the parameters of the model. The main goal of OSSL is to learn an ID classification model described by the parameter $\theta$ optimized by minimizing $\mathcal{L}$ on $\mathcal{S}$ and $\U$. And the most tricky problem is how to handle the open-set data set $\U$ so that better ID classification performance can be gained.

Previous studies have presented different answers to this tricky problem. Taking OpenMatch~\cite{li2023iomatch} as an example, given a labeled example $\xl_{i}$, it will acquire a $K$-dimensional probability vector $\p(\theta;\xl_i)$ by a traditional ID classifier and $K$ $2$-dimensional probability vector $\{\q^{k}(\theta;\xl_i)=(q^{k}_0(\theta;\xl_i), q^{k}_1(\theta;\xl_i))\}_{k=1}^{K}$ by an OOD detector which is composed of $K$ binary sub-classifier. Here $q^{k}_0(\theta;\xl_i)$ indicates the probability of being an inlier of class $k$ while $q^{k}_1(\theta;\xl_i)$ indicates not. For simplicity, we will hide the model's parameters $\theta$ and reduce the notation $\p(\theta;\xl_i)$ and $\q^{k}(\theta;\xl_i)=(q^{k}_0(\theta;\xl_i), q^{k}_1(\theta;\xl_i))$ to $\p(\xl_i)$ and $\q^{k}(\xl_i)=(q^{k}_0(\xl_i), q^{k}_1(\xl_i))$, respectively. Then to train the model upon labeled training data set $\mathcal{S}$, OpenMatch will compute the following losses:
\begin{equation}
    \label{eqn:loss:ce}
    \mathcal{L}_{ce}(\theta;\mathcal{S})=\frac{1}{|\mathcal{S}|} \sum_{(\xl_i,\y_i) \in \mathcal{S}} H(\y_{i}, \p(\xl_i)),
\end{equation}
\begin{equation}
    \label{eqn:loss:ova}
    \mathcal{L}_{ova}(\theta;\mathcal{S})=
    -\frac{1}{|\mathcal{S}|} \sum_{(\xl_i,\y_i) \in \mathcal{S}}
    \log{q^{y_i}_0(\xl_i)}+ \mathop{\min}\limits_{k\neq y_{i}}\log{q^{k}_1(\xl_i)},
\end{equation}
where $H(\cdot,\cdot)$ denotes the standard cross-entropy loss, $|\mathcal{S}|=N_l$ denotes the cardinality of $S$, and $y_{i}$ denotes the ground truth of $\x_i^{l}$, \emph{i.e.}, $\x_i^{l}$ belongs to class $y_{i}$. On the other hand, given an unlabeled instance $\xu_i$, OpenMatch first applies standard random cropping $\alpha(\cdot)$ as weak data augmentation to obtain $\alpha_{0}(\xu_i)$ and $\alpha_{1}(\xu_i)$. Then it will calculate the following losses:
\begin{equation}
    \label{eqn:loss:em}
    \mathcal{L}_{em}(\theta;\U)= -\frac{1}{|\U|} \sum_{\xu_i\in\U} \sum_{j=0}^{1} \sum_{k=1}^{K} \q^{k}(\alpha_{j}(\xu_i))\log{\q^{k}(\alpha_{j}(\xu_i))},
\end{equation}
\begin{equation}
    \label{eqn:loss:oc}
    \mathcal{L}_{oc}(\theta;\U)= \frac{1}{|\U|} \sum_{\xu_i\in\U} \sum_{k=1}^{K} \|\q^{k}(\alpha_{0}(\xu_i))-\q^{k}(\alpha_{1}(\xu_i))\|_2^2.
\end{equation}
Moreover, it adopts FixMatch over the pseudo-ID instances which can be formulated as
\begin{equation}
    \label{eqn:loss:fm}
    \mathcal{L}_{fm}(\theta;\U) =
    \frac{1}{|\U|} \sum_{\xu_i\in\U} \mathcal{M}(\alpha(\xu_i)) H(\hat{\y}_i^u, \p(\mathcal{A}(\xu_i))),
\end{equation}
where $\mathcal{A}(\cdot)$ indicates strong data augmentation like RandAugment~\cite{cubuk2020randaugment} and $\mathcal{M}(\alpha(\xu_i)) =  \mathbb{I}(q^{\hat{y}_i^u}_0(\alpha(\xu_i))>0.5) \cdot \mathbb{I}(\max(\p(\alpha(\xu_i)))>\rho)$ in which $\rho$ is a threshold, and $\hat{y}_i^u=\arg\min \p(\alpha(\xu_i))$ represents the pseudo-label of $\xu_i$ and $\hat{\y}_i^u$ will be its one-hot version. To summarize, the overall objective function of OpenMatch can be written as
\begin{align}
    \label{eqn:loss:op}
    \nonumber \mathcal{L} = & \underbrace{\mathcal{L}_{ce}(\theta;\mathcal{S}) + \mathcal{L}_{ova}(\theta;\mathcal{S})}_{\mathcal{L}_s(\theta;\mathcal{S})}  \\
    &+\underbrace{\lambda_1\mathcal{L}_{em}(\theta;\U) + \lambda_2\mathcal{L}_{oc}(\theta;\U) + \lambda_3\mathcal{L}_{fm}(\theta;\U)}_{\mathcal{L}_u(\theta;\U)},
\end{align}
where $\lambda_1$, $\lambda_2$ and $\lambda_3$ are trade-off parameters. It can be observed that OpenMatch recklessly employs the whole open-set data set $\U$ to train the model which is common in previous studies. Note that although in Eq.\ref{eqn:loss:fm}, $\mathcal{M}(\alpha(\xu_i))$ is applied for obtaining confident pseudo-ID instances, Eq.\ref{eqn:loss:em} and Eq.\ref{eqn:loss:oc} still utilize the whole unlabeled data set. Thus, the entire model of OpenMatch actually still leverages all open-set data. Oppositely, we argue that it is necessary to selectively leverage the open-set data $\U$ which will be demonstrated in the following theoretical analysis.

\subsection{Theoretical Understanding}\label{sec:thm:und}
In this subsection, we aim to understand the learning task from the perspective of generalization. For the simplicity of generalization analysis, we abstract the key points of the learning task and make it more math-friendly. For example, we do not consider data augmentations in our analysis. To this end, we formulate it as the following risk minimization (RM) problem, which is commonly used in Statistical Learning Theory: 
\begin{align}\label{prob:rm}
    \min_{\theta}\Lmath(\theta):=\E_{\zeta\sim\mathcal{D}}[\ell(\theta;\zeta)],
\end{align}
where $\theta$ is the parameter to be learned, $\zeta$ is the training data following a unknown distribution $\mathcal{D}$, $\ell$ is the loss function and $\E[\cdot]$ is the expectation. Generally, it is impossible to know the loss function $\Lmath(\theta)$ explicitly due to the unknown distribution of $\mathcal{D}$. Instead of solving problem~(\ref{prob:rm}) directly, we usually consider the following empirical risk minimization (ERM) problem:
    $\min_{\theta}\Lt(\theta):=\frac{1}{|\Dt|}\sum\limits_{\zeta\in\Dt}\ell(\theta;\zeta)$,
where $\Dt$ is a sampled training data set with sample size $|\Dt|$. To solve the ERM problem, stochastic gradient descent (SGD) is widely employed, whose key update step is $\theta_{t+1} = \theta_t - \eta \gt(\theta_t)$ with the learning rate $\eta>0$, where $\theta_t$ indicates the parameter in $t$-th epoch. Due to the labeled (ID) and unlabeled (ID and OOD) data containing in $\Dt$, the stochastic gradient of loss function computed at $\theta_t$ can be rewritten as 
\begin{align}
    \gt(\theta_t) = \lambda \gi(\theta_t) + (1-\lambda) (\tau \gn(\theta_t) + (1-\tau)\gf(\theta_t)),
\end{align}
where $\lambda,\tau \in[0,1]$ are two constants, $\gi$ is the stochastic gradient computed by ID data while $\gn$ and $\gf$ are two stochastic gradients computed by friendly unlabeled data and unfriendly unlabeled data respectively. We use the gradient variances to measure the distance between labeled data and open-set data. Specifically, we bound the variance for each stochastic gradient in the following assumptions.
\begin{assumption}[Bounded variance~\cite{ghadimi2013stochastic}]\label{ass:grad}
The stochastic gradient is unbiased, $\E[ \gt(\theta)] =\nabla \Lmath(\theta)$.
The stochastic gradient $\gi(\theta)$ is variance bounded, i.e., there exists a constant $\sigma^2>0$, such that
\begin{align*}
\E[\|\gi(\theta)-\nabla\Lmath(\theta)\|^2] \le \sigma^2. 
\end{align*}
\end{assumption}
\begin{assumption}[Weak Growth Condition~\cite{bertsekas1995neuro, bottou2018optimization}]\label{ass:grad:Fb}
The stochastic gradient of $\gn(\theta)$  and $\gf(\theta)$  are  variance bounded, i.e., there exists a constant $\sigma^2>0$, such that
\begin{align*}
&\E[\|\gn(\theta)-\nabla\Lmath(\theta)\|^2] \le  \frac{\epsilon}{2}\|\nabla\Lmath(\theta)\|^2 + \sigma^2.\\
    &\E[\|\gf(\theta)-\nabla\Lmath(\theta)\|^2] \le \frac{\nu}{2}\|\nabla\Lmath(\theta)\|^2 + \sigma^2.
\end{align*}
where $\epsilon>0$ is a small constant and $\nu \gg 1$ is a large enough constant.
\end{assumption}
Since $\nu \gg 1$ is large enough, we can consider that the variance for $\gf(\theta)$ is much larger than the variance for $\gn(\theta)$. 
We consider the friendly data to have small variance so we suppose $\epsilon > 0$ is small. Similarly, we consider $\nu \gg 1$ is large enough for unfriendly data. In generalization analysis, we are interested in the excess risk bound (ERB):
\begin{align}
    \text{ERB}(\widehat\theta) := \Lmath(\widehat\theta) - \Lmath(\theta_*),
\end{align}
where $\widehat \theta$ is a solution obtained by an algorithm and $\theta_*\in\arg\min_{\theta}\Lmath(\theta)$ is the optimal solution of problem (\ref{prob:rm}). For the convenience of analysis, we make the following widely used assumptions for the loss function. 
\begin{assumption}[Smoothness~\cite{opac-b1104789}]\label{ass:smooth}
$\Lmath(\theta)$ is smooth with an $L$-Lipchitz continuous gradient, i.e., it is differentiable and there exists a constant $L>0$ such that 
\begin{align*}
    \|\nabla \Lmath(\theta)  - \nabla \Lmath(\theta')\|\leq L\|\theta - \theta'\|.
\end{align*}
This is equivalent to
\begin{align*}
    \Lmath(\theta) - \Lmath(\theta') \le \langle \Lmath(\theta'), \theta - \theta' \rangle + \frac{L}{2}\|\theta-\theta'\|^2.
\end{align*} 
\end{assumption}

\begin{assumption}[Polyak-\L ojasiewicz condition~\cite{polyak1963gradient}]\label{ass:PL}
There exists a constant $\mu>0$ such that \begin{align*}
2\mu (\Lmath(\theta) - \Lmath(\theta_*)) \le \|\nabla \Lmath(\theta)\|^2,
\end{align*}
where $\theta_* \in\arg\min_{\theta} \Lmath(\theta)$ is a optimal solution.
\end{assumption}
\jn{It is worth noting that the Polyak-\L ojasiewicz condition has been theoretically~\cite{allen2019convergence} and empirically~\cite{Yuan0JY19} observed in training deep neural networks. Weaker than many other conditions like strong convexity, restricted strong convexity, and weak strong convexity~\cite{KarimiNS16}, it has been widely used to establish the convergence of non-convex optimization~\cite{abs-2002-05273,CharlesP18,LiL18}.}

Under the standard assumptions on loss function, we have the following theorem for ERB. Due to the space limitation, we include the proof in the supplementary.
\begin{thm}\label{thm:1}
Under assumptions~\ref{ass:grad}, \ref{ass:grad:Fb}, \ref{ass:smooth}, \ref{ass:PL}, we have the following ERB in expectation: \\
(a) when all data are used: by setting $\eta \le \frac{1}{(1-\lambda)(\tau\epsilon + (1-\tau)\nu) L}$, we have $$\text{ERB}(\widehat\theta_{\text{id+uf+fr}}) \le O\left(\Lmath(\theta_0)-\Lmath(\theta_*)\right);$$ 
(b) when only labeled data are used: by setting $\eta = \frac{2}{n\mu} \log\left( \frac{n\mu^2(\Lmath(\theta_0)-\Lmath(\theta_*))}{\sigma^2L}\right)$, we have 
\begin{align*}
    \text{ERB}(\widehat\theta_{\text{id}}) &\le \frac{L\sigma^2}{n\mu^2}+ \frac{2L\sigma^2}{n\mu^2} \log\left( \frac{n\mu^2(\Lmath(\theta_0)-\Lmath(\theta_*))}{\sigma^2L}\right)\\
    &\le  O\left( \frac{\log\left( n \right)}{n}  \right);
\end{align*}
(c) when labeled and friendly data are used: by setting $\eta = \frac{2}{(n+m)\mu} \log\left( \frac{(n+m)\mu^2(\Lmath(\theta_0)-\Lmath(\theta_*))}{\sigma^2L}\right)$, we have
\begin{align*}
    \text{ERB}(\widehat\theta_{\text{id+fr}}) &\le \frac{L\sigma^2}{(n+m)\mu^2}\\
    &\ \ \ +\frac{2L\sigma^2}{(n+m)\mu^2} \log\left( \frac{(n+m)\mu^2(\Lmath(\theta_0)-\Lmath(\theta_*))}{\sigma^2L}\right)\\
    &\le O\left( \frac{\log\left( n+m \right)}{(n+m)}  \right),
\end{align*}
where $n$ and $m$ are the sample sizes of labeled data and friendly data, respectively, $ \widetilde O (\cdot)$ suppresses a logarithmic factor and constants.
\end{thm}

The results of Theorem~\ref{thm:1} show that (1) the algorithm could not reduce the objective due to the large variance arising from unfriendly open-set data; (2) by using friendly open-set data, the algorithm could significantly reduce the objective, and it has better generalization by comparing with the one only using labeled data. The theoretical findings inspire us to design a selection method for wisely leveraging open-set data. Specifically, one may carefully select and use friendly open-set data during training progress to improve the performance of the learning task.

\subsection{Wise Selection Mechanism}
\label{sec:sel}

Inspired by the theoretical understanding above, we aim to wisely select and exploit a subset of the original unlabeled data set $\U_{t}\subseteq \U$ that consists of the open-set data friendly to the model training in the $t$-th epoch so that we can learn a model from $\U_{t}$ and $\mathcal{S}$ with better capability of classifying ID instances. 

To this end, based on Theorem~\ref{thm:1}, we design a gradient-variance-based selection mechanism (GV-SM) to discard the unfriendly open-set data with large gradient variance so that we can exploit the remaining relatively friendly open-set data to learn a model with better ID classification capability. Specifically, our proposed GV-SM of the $t$-th epoch can be formulated as 
\begin{equation}
    \label{eqn:gvsm}
    \U_{t}=
    \left\{
        \xu_i\in\U\ |\ \|\gxu(\theta_{t}) - \bar{g}(\theta_{t})\| < \sqrt{\rho_{t}}
    \right\},
\end{equation}
where $\gxu(\theta_{t})$ denotes the gradient computed by open-set instances $\xu_i$, $\bar{g}(\theta_{t})$ denotes the estimated expectation of the gradient of the overall objective function $\mathcal{L}$, and $\rho_{t}$ indicates the threshold in the $t$-th epoch. In this paper, we utilize the labeled data set $\mathcal{S}$ to obtain $\bar{g}(\theta_{t})$: 
\begin{equation}
    \label{eqn:g_bar}
    \bar{g}(\theta_{t}) = \frac{1}{N_{l}} \sum_{i=1}^{N_{l}} \frac{\partial \mathcal{L}_{s}(\theta_t; \xl_i)}{\partial \theta_{t}}.
\end{equation}
Meanwhile, we calculate $\gxu(\theta_{t})$  by the following formula:
\begin{equation}
    \label{eqn:g_xu}
    \gxu(\theta_{t}) = \frac{\partial \mathcal{L}_{u}(\theta_t;\xu_i)}{\partial \theta_{t}}.
\end{equation}
As for obtaining $\rho_{t}$, without loss of generality, we apply the following two simple methods: (1) Top-k, which utilizes the $k$-th largest gradient variance among $\{\gxu(\theta_{t})\}_{i}^{N_{u}}$ as $\rho_{t}$; (2) Otsu thresholding~\cite{otsu1979threshold}, which adaptively determine $\rho_{t}$ by maximizing the variance of $\gxu(\theta_{t})$ between the selected and discarded open-set data clusters. 

By applying this wise selection mechanism, WiseOpen reformulates the typical OSSL objective function in the $t$-th epoch as
\begin{equation}
    \label{eqn:loss:overall_new}
    \mathcal{L} = \mathcal{L}_{s}(\theta_t; \mathcal{S}) + \mathcal{L}_{u}(\theta_t;\U_t).
\end{equation}
To be more specific, after implementing our proposed WiseOpen on top of OpenMatch, we will rewrite OpenMatch's objective formulated in Eq.\ref{eqn:loss:op} as
\begin{align}
    \label{eqn:loss:newop}
    \nonumber \mathcal{L} = &\underbrace{\mathcal{L}_{ce}(\theta_t;\mathcal{S}) + \mathcal{L}_{ova}(\theta_t;\mathcal{S})}_{\mathcal{L}_s(\theta_t;\mathcal{S})} \\
    &+ \underbrace{\lambda_1\mathcal{L}_{em}(\theta_t;\U_t) + \lambda_2\mathcal{L}_{oc}(\theta_t;\U_t) + \lambda_3\mathcal{L}_{fm}(\theta_t;\U_t)}_{\mathcal{L}_u(\theta_t;\U_t)}.
\end{align}
\begin{algorithm}[t]
    \caption{WiseOpen Family.}
    \label{alg:main}
    \KwIn{Labeled data $\mathcal{S}$, unlabeled data $\mathcal {U}$, model parameters $\theta$, epoch $E_{max}$, iteration $I_{max}$, learning rate $\eta$, selection interval $e_s$.}
    \For {$t \leftarrow 1\ to\ E_{max}$} {

        \eIf{$t$ \% $e_s$ == 0}{
            \textbf{Obtain} $\U_{t}$ according to Eq.\ref{eqn:gvsm} or Eq.\ref{eqn:lsm};
        }{
            \textbf{Obtain} $\U_{t} = \U_{t-1}$;
        }
     
        \For {$iter \leftarrow1\ to\ I_{max}$} {
            \textbf{Sample} batches $\mathcal{B}_{l}\in\mathcal{S}$ and $\mathcal{B}_{u}\in\mathcal{U}_{t}$;
            
            \textbf{Compute} $\mathcal{L}\leftarrow\mathcal{L}_{s}(\theta;\mathcal{B}_{l}) + \mathcal{L}_{u}(\theta;\mathcal{B}_{u})$;

            \textbf{Update} $\theta \leftarrow \theta - \eta\frac{\partial\mathcal{L}}{\partial\theta }$;           
        }
        
    } 
\end{algorithm}

\subsection{Practical Variants}
Ideally, the selection procedure should be implemented for every epoch, which is applied in the naive WiseOpen. However, it can be computationally expensive to calculate gradient variance for each unlabeled instance in every epoch, as shown in Table~\ref{tb:runing_time} in Section~\ref{results}. Therefore, we proposed two practical variants of WiseOpen, namely WiseOpen-E and WiseOpen-L. The overall framework of the WiseOpen family, \emph{i.e.}, WiseOpen and its variants, is summarized in Algorithm~\ref{alg:main}.

\textbf{WiseOpen-E.} Before introducing WiseOpen-E, let us consider an example as follows. The updating step of SGD at $t$-th epoch can be formulated as $\theta_{t} = \theta_{t-1} - \eta g(\theta_{t-1})$, then after $m$ epochs, we have
$\theta_{t+m} = \theta_{t-1} - \eta \sum_{k=0}^{m} g(\theta_{t-1+k})$. By the condition of smoothness variance with parameter $L'$, we have
\begin{align}
    \|g(\theta_{t+m}) - g(\theta_t)\| 
    \le L'\| \theta_{t+m} - \theta_{t}\|
    = \eta L' \left\|\sum_{k=1}^{m} g(\theta_{t-1+k})\right\|.
\end{align}
Since $\eta$ is small, if $m$ is not large, then $\eta L' \left\|\sum_{k=1}^{m} g(\theta_{t-1+k})\right\|$ is small, indicating that $g(\theta_{t+m})$ and $ g(\theta_t)$ are close enough. Inspired by this example, gradients within a small updating interval could be similar. Thus, as a natural idea, an economical version of WiseOpen, WiseOpen-E simply sets an interval $e_s$ of updating the selecting result of open-set data. Specifically, once a selection procedure in the $t$-th epoch is accomplished, the selected open-set subset will remain unchanged for $(e_s-1)$ epochs, \emph{i.e.}, $\U_{t} = \U_{t+1}=\dots=\U_{t+e_s-1}$, and in the $(t+e_s)$-th epoch another selection procedure will be carried out to update the selected open-set subset. In our experiments of WiseOpen-E, $e_s$ is set as $10$ while for the experiments of WiseOpen and WiseOpen-L introduced later, $e_s$ is set as $1$.
Nevertheless, as a trade-off of obtaining higher efficiency, the stale selection issue will inevitably arise in WiseOpen-E. This means the selected subset may be outdated for the current model training since the selection decision is based on the previous, older model.

\begin{table*}[!htbp]
    \begin{center}
    \caption{The hyper-parameter $k$ for the Top-k threshold.}
    \label{tb:topk}

    \subtable[$k$ utilized in the Top-k threshold for GV-SM followed with the proportion to the whole open-set.]{
        \begin{tabular}{l c ccc c cc c c}
            \toprule
             && \multicolumn{3}{c}{CIFAR-10} && \multicolumn{2}{c}{CIFAR-100} && \multicolumn{1}{c}{Tiny-ImageNet} \\
            \cmidrule{3-5} \cmidrule{7-8}  \cmidrule{10-10} 
            Vanilla Baselines && 50 labels & 100 labels & 400 labels && 50 labels & 100 labels  && 50 labels\\
            \midrule
            \noalign{\smallskip}
            FixMatch
            && 2470 (5\%) & 2455 (5\%)  & 2365 (5\%)
            && 880  (2\%) & 410  (1\%)
            && 1760 (2\%)\\
            FreeMatch
            &&  2470 (5\%) & 2455 (5\%)  & 4730 (10\%)
            &&  6600 (15\%) &  4100(10\%)
            &&  17600 (20\%) \\
            MTC
            && 4940 (10\%) & 4910 (10\%)& 2365 (5\%)
            && 4400 (10\%) & 4100 (10\%)
            && 4400 (5\%)\\
            OpenMatch
            && 2470 (5\%)  & 2455 (5\%) & 2365 (5\%)
            && 4400 (10\%) & 4100 (10\%)
            && 4400 (5\%)\\
            IOMatch
            && 2470 (5\%)  & 982 (2\%)  & 946 (2\%)
            && 880 (2\%)   & 2050 (5\%)
            && 1760 (2\%)\\
            \bottomrule
        \end{tabular}
    }

    \subtable[$k$ utilized in the Top-k threshold for L-SM followed with the proportion to the whole open-set.]{
        \begin{tabular}{l c ccc c cc c c}
            \toprule
             && \multicolumn{3}{c}{CIFAR-10} && \multicolumn{2}{c}{CIFAR-100} && \multicolumn{1}{c}{Tiny-ImageNet} \\
            \cmidrule{3-5} \cmidrule{7-8}  \cmidrule{10-10} 
            Vanilla Baselines && 50 labels & 100 labels & 400 labels && 50 labels & 100 labels  && 50 labels\\
            \midrule
            \noalign{\smallskip} 
            MTC
            && 4940 (10\%) & 4910 (10\%)& 2365 (5\%)
            && 4400 (10\%) & 4100 (10\%)
            && 4400 (5\%)\\
            OpenMatch
            && 2470 (5\%)  & 2455 (5\%) & 2365 (5\%)
            && 4400 (10\%) & 4100 (10\%)
            && 4400 (5\%)\\
            IOMatch
            && 494   (1\%)   & 491  (1\%)  & 473 (1\%)
            && 6600  (15\%)  & 6150 (15\%)
            && 22000 (25\%)\\
            \bottomrule
        \end{tabular}
    }
    \end{center}
\end{table*}

\textbf{WiseOpen-L.} Considering in practice, calculating the loss value for each sample is significantly less computationally expensive compared to calculating the gradient variance, if loss value can be employed as a substitute for the gradient variance, then we can achieve computationally friendly data selection without raising the stale selection issue. Inspired by the Polyak-\L ojasiewicz condition~\cite{polyak1963gradient} of a loss function $\ell(\theta)$:
\begin{align}
    \ell(\theta) \le \frac{1}{2\mu}\|\nabla \ell(\theta)\|^2 + \ell(\theta_*),
\end{align}
where $\mu>0$ is a constant and $\theta_* \in\arg\min_{\theta} \ell(\theta)$ is a optimal solution, we can make an informal connection between loss function and its gradient norm. 
If we apply the Polyak-\L ojasiewicz condition to function $\mathcal{L}_{u}(\theta; \xu_i)$, then we have
\begin{align}
     \nonumber \mathcal{L}_{u}(\theta_t; \xu_i)  \le& \frac{1}{2\mu}\|\gxu(\theta_t)\|^2 + \mathcal{L}_{u}(\theta_*; \xu_i).
\end{align}
Once $\|\gxu(\theta_t) - \bar{g}(\theta_t)\| \le \sqrt{\rho_t}$ holds in (\ref{eqn:gvsm}), then by $\|\gxu(\theta_t) - \bar{g}(\theta_t)\| \ge \|\gxu(\theta_t) \| - \|\bar{g}(\theta_t)\|$ we have $\|\gxu(\theta_t) \|  \le \sqrt{\rho_t} + \|\bar{g}(\theta_t)\|$, so that
\begin{align}
\mathcal{L}_{u}(\theta_t; \xu_i) \le \frac{(\sqrt{\rho_t} + \|\bar{g}(\theta_t)\|)^2}{2\mu}+ \mathcal{L}_{u}(\theta_*; \xu_i) .
\end{align}
That is to say, if $\xu_i\in\mathcal{U}_t$ in (\ref{eqn:gvsm}), i.e. $\|\gxu(\theta_t) - \bar{g}(\theta_t)\|$ is upper bounded, then $\mathcal{L}_{u}(\theta_t; \xu_i)$ can aslo be upper bounded.
Therefore, we propose another variant of WiseOpen, called WiseOpen-L which applies a loss-based selection mechanism (L-SM) that selects the friendly open-set data with smaller loss values to construct $\U_{t}$: 
\begin{equation}
    \label{eqn:lsm}
    \U_{t}=
    \left\{
        \xu_i\in\U\ |\ \mathcal{L}_{u}(\theta_t;\xu_i) < \rho'_{t}
    \right\},
\end{equation}
where in this paper, the threshold $\rho'_{t}$ is also acquired by the Top-k or Otsu thresholding methods.

\subsection{Relationships to Previous Methods}
\label{sec:relation}
\textbf{Inclusiveness to Previous Methods.} Firstly, our proposed WiseOpen and its variants are designed upon the unified formulation of the overall objectives in the OSSL scenarios. Therefore, we are proposing robust generic OSSL frameworks that can be easily implemented into the existing OSSL and SSL methods to improve performance. Moreover, WiseOpen-L, as one of our proposed accelerated variants, actually can be seen as a general version of some previous SSL and OSSL methods that adopt confidence-based selection mechanisms. For example, when we exclude $\mathbb{I}(q^{\hat{y}_i^u}(\alpha(\xu_i))>0.5)$ in $\mathcal{M}(\alpha(\xu_i))$ of $\mathcal{L}_{fm}$ formulated in Eq.\ref{eqn:loss:fm}, it will be the exact unsupervised loss $\mathcal{L}_{u}$ in FixMatch~\cite{SohnBCZZRCKL20}. And the remaining selecting component $\mathbb{I}(\max(\p(\alpha(\xu_i)))>\rho)$ can converted to $\mathbb{I}(-\log{\max(\p(\alpha(\xu_i)))}<-\log{\rho})$, which actually can be seen as a kind of loss-based selection mechanism depending on the cross-entropy loss over $\alpha(\xu_i)$ using its one-hot pseudo-label.

\textbf{Differnce between Exisiting Selection Mechanisms}. Although previous studies have exploited various selection mechanisms, we distinguish our proposed method in the following aspects: (1) Our frameworks apply the selection mechanism on the whole learning task level, or in other words, on the unified unsupervised loss level. In contrast, the previous methods usually employ selection mechanisms on a single loss function level. For example, in OpenMatch, $\mathcal{L}_{fm}$ formulated in Eq.\ref{eqn:loss:fm} uses $0.5$ and $\rho$ as the threshold to select the pseudo-ID instances with high confidence in ID classification. But this selection is just applied to this single loss while $\mathcal{L}_{em}$ and $\mathcal{L}_{oc}$ formulated in Eq.\ref{eqn:loss:em} and Eq.\ref{eqn:loss:oc} still leverage the whole open-set data, which may harm the model's performance due to the existence of the unfriendly open-set data. (2) As far as we know, different from the popular confidence-based selection mechanisms used in the previous work~\cite{SohnBCZZRCKL20,li2023iomatch,saito2021openmatch,Wang0HHFW0SSRS023}, we are the first to propose selecting open-set data based on the gradient variance with solid theoretical analysis in the OSSL scenarios.

\section{Experiments}
\label{expe}

In this section, we introduce the comprehensive settings of experiments and provide sufficient evaluations to demonstrate the effectiveness of our proposed frameworks.

\subsection{Experimental Settings}
\label{expe:settings}

\textbf{Datasets.} Following~\cite{CaoBL22, guorobust, onpseudo2023, saito2021openmatch, yu2020multi}, we choose three benchmark datasets to evaluate the efficacy of WiseOpen, namely: 
\begin{itemize}
    \item CIFAR-10~\cite{krizhevsky2009learning}, a dataset consisting of 10 classes, of which each class contains 5,000 and 1,000 images for training and testing respectively;
    \item CIFAR-100, a dataset consisting of 100 classes, of which each class contains 500 and 100 images for training and testing respectively;
    \item Tiny-ImageNet~\cite{yao2015tiny}, a subset of ImageNet~\cite{DengDSLL009} consisting of 200 classes, of which each class contains 500 and 100 images for training and testing respectively.
\end{itemize}
For all datasets, different from~\cite{saito2021openmatch, li2023iomatch}, the first $K$ classes are taken as seen classes while the rest of classes are taken as unseen classes, in order to construct a more complicated OSSL scenario followed by~\cite{CaoBL22, guorobust}. Following~\cite{saito2021openmatch}, we conduct the experiments with different amounts of labeled training examples: 50, 100, or 400 labels per seen class for CIFAR-10, and 50 or 100 labels per seen class for CIFAR-100. As for Tiny-ImageNet, 50 training examples are taken as labeled data for each seen class. Additionally, 50 training examples per seen class are split as a validation set for all experiments. Except for the labeled set and validation set, the remaining training data are taken as unlabeled instances.

\begin{figure}[!t]
    \subfigure[Evaluation of ID classification.]{
        \centering
        \includegraphics[width=\textwidth]{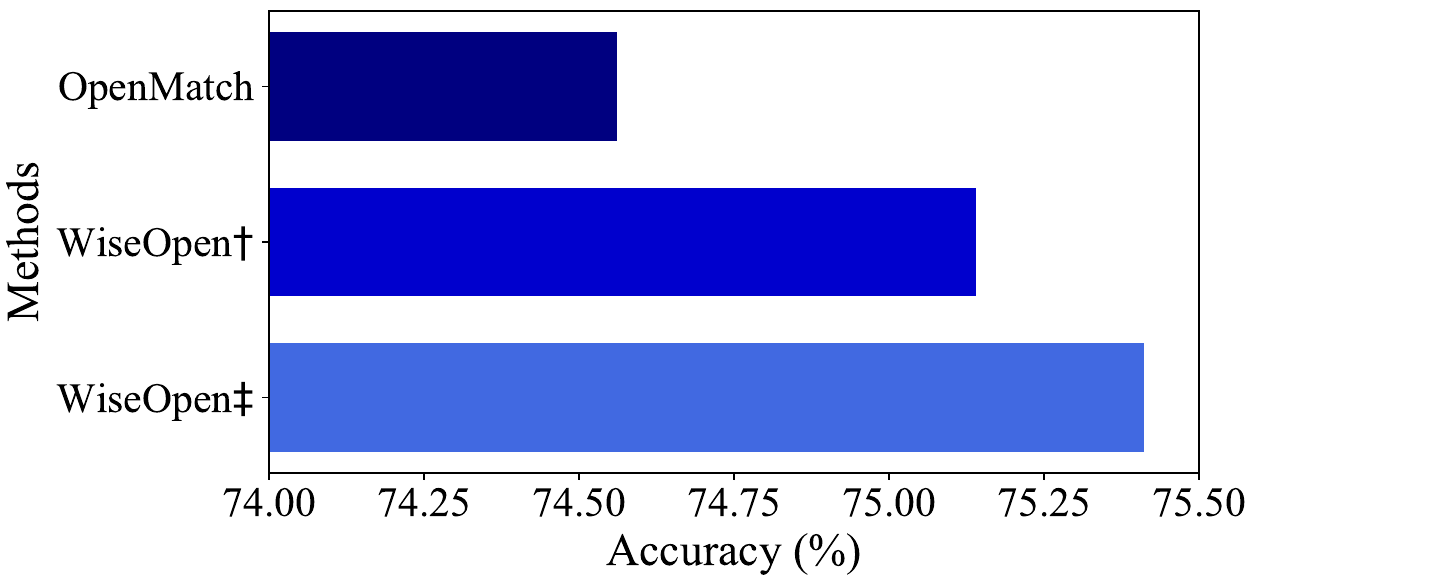}
    }
    
    \subfigure[Evaluation of OOD detection.]{
        \centering
        \includegraphics[width=\textwidth]{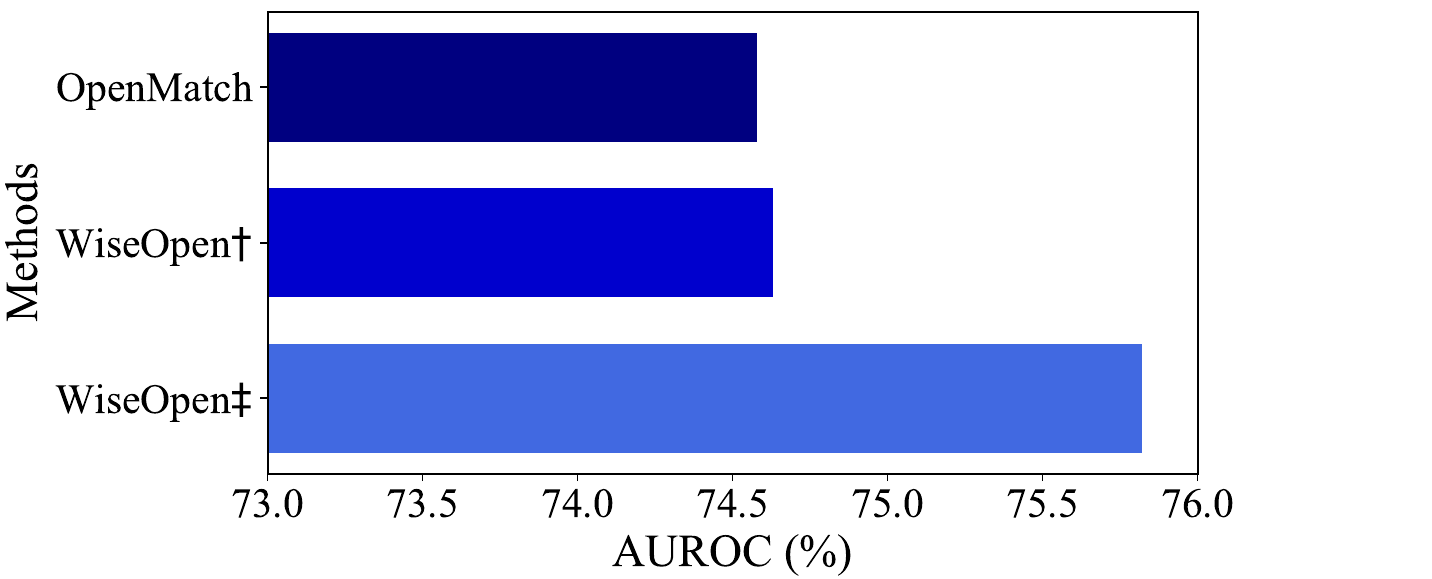}
    }
    \caption{Performance of original OpenMatch and our proposed WiseOpen on top of OpenMatch. Experiments are conducted on CIFAR-100 with 100 labels per class. $\dagger$ means using Top-k threshold while $\ddagger$ means using Otsu threshold.}
    \label{fig:wiseopen}
    \vspace{-2mm}
\end{figure}

\begin{table}[!t]
    \centering
    \begin{tabular}{lc}
        \toprule
        Algorithm & Training Time \\
        \midrule
        OpenMatch    &  10h 59m \\
        ~~w/ WiseOpen $\dagger$    &  43h 01m \\
        ~~w/ WiseOpen $\ddagger$   &  43h 05m \\
        ~~w/ WiseOpen-E $\dagger$  &  13h 40m \\
        ~~w/ WiseOpen-E $\ddagger$  &  13h 43m \\
        ~~w/ WiseOpen-L $\dagger$ &  11h 36m\\
        ~~w/ WiseOpen-L $\ddagger$ &  11h 43m\\
        \bottomrule
    \end{tabular}
    \caption{Models' training time of original OpenMatch and our proposed frameworks on top of OpenMatch. Experiments are conducted using CIFAR-100 with 100 labels on a single  NVIDIA GeForce RTX 4090.}
    \label{tb:runing_time}
\end{table}

\begin{table*}[!t]
    \centering
    \caption{Comparison of ID classification accuracy (in \%, mean $\pm$ standard deviation) on CIFAR-10, CIFAR-100, and Tiny-ImageNet with varying labels per seen class. 
    $\dagger$ means using Top-k threshold while $\ddagger$ means using Otsu threshold. 
    Each block consists of the results of the baseline with or without the variants of WiseOpen and the improvements that our proposed frameworks can make. The best results for each data setting in each block are in bold.}
    \label{tb:main_results:acc}
    \begin{tabular}{l c ccc c cc c c}
        \toprule
         && \multicolumn{3}{c}{CIFAR-10} && \multicolumn{2}{c}{CIFAR-100} && \multicolumn{1}{c}{Tiny-ImageNet} \\
        \cmidrule{3-5} \cmidrule{7-8}  \cmidrule{10-10} 
        Algorithms && 50 labels & 100 labels & 400 labels && 50 labels & 100 labels  && 50 labels\\
        \midrule
        \noalign{\smallskip}
        Labeled Only
        && 63.16$\pm$0.87 & 67.26$\pm$1.08 & 83.67$\pm$0.29 
        && 60.15$\pm$0.20 & 64.96$\pm$0.29 
        && 42.31$\pm$0.43\\
        \midrule
        FixMatch~\cite{SohnBCZZRCKL20} 
        && 90.48$\pm$0.01 & \bf{92.61$\pm$0.16} & {93.68$\pm$0.46} 
        && {70.16$\pm$0.48} & \bf{74.17$\pm$0.22} 
        && \bf{45.11$\pm$0.53}\\
        ~~w/ WiseOpen-E $\dagger$
        && {91.33$\pm$0.15} & \bf{92.61$\pm$0.32} & 93.67$\pm$0.20
        && \bf{70.53$\pm$0.48} & {74.13$\pm$0.41}
        && {44.90$\pm$0.63}\\
        ~~w/ WiseOpen-E $\ddagger$
        && \bf{91.52$\pm$0.44} & 92.54$\pm$0.11 & \bf{93.80$\pm$0.31}
        && 69.86$\pm$0.16 & 74.12$\pm$0.38
        && 44.98$\pm$0.57\\
       ~~$\Delta$ (mean)
        && +0.94 & -0.03 & +0.06
        && +0.04 & -0.04
        && -0.17\\
        ~~$\Delta$ (max)
        && +1.04 & 0.00 & +0.12
        && +0.38 & -0.04
        && -0.13\\
        \midrule
        FreeMatch~\cite{Wang0HHFW0SSRS023} 
        && 85.43$\pm$0.64 & 88.44$\pm$0.36 & 89.47$\pm$0.29 
        && 65.38$\pm$0.80 & 70.20$\pm$0.27
        && 42.16$\pm$0.84\\
        ~~w/ WiseOpen-E $\dagger$
        && \bf{86.09$\pm$2.00} & {88.36$\pm$0.80} & 89.76$\pm$0.75
        && \bf{65.68$\pm$0.52} & \bf{70.32$\pm$0.40}
        && \bf{42.74$\pm$0.06}\\
        ~~w/ WiseOpen-E $\ddagger$
        && {85.71$\pm$0.33} & \bf{88.71$\pm$0.36} & \bf{90.37$\pm$0.76}
        && 65.52$\pm$0.47 & 70.13$\pm$0.12
        && 42.49$\pm$0.55\\
        ~~$\Delta$ (mean)
        && +0.47 & +0.09 & +0.60
        && +0.22 & +0.02
        && +0.46\\
        ~~$\Delta$ (max)
        && +0.66 & +0.27 & +0.90
        && +0.30 & +0.12
        && +0.58\\
        \midrule
        MTC~\cite{yu2020multi}
        && 79.00$\pm$1.73 & 80.51$\pm$1.67 & 89.03$\pm$0.93 
        && 64.22$\pm$0.61 & 70.22$\pm$0.57
        && 39.57$\pm$0.17\\
        ~~w/ WiseOpen-E $\dagger$
        && 81.37$\pm$2.71 & 82.58$\pm$1.51 & \bf{89.73$\pm$0.34}
        && \bf{64.71$\pm$0.28} & 70.33$\pm$0.12
        && \bf{40.49$\pm$0.48}\\
        ~~w/ WiseOpen-E $\ddagger$
        && {82.57$\pm$0.40} & 82.17$\pm$1.08 & 89.27$\pm$0.45 
        && 64.54$\pm$0.48 & \bf{70.42$\pm$0.16}
        && 39.80$\pm$0.11\\
        ~~w/ WiseOpen-L $\dagger$
        && 81.34$\pm$1.89 & {83.45$\pm$1.45} & 89.23$\pm$0.87 
        && {64.68$\pm$0.83} & 70.16$\pm$0.76
        && {39.83$\pm$0.28}\\
        ~~w/ WiseOpen-L $\ddagger$
        && \bf{82.65$\pm$0.32} & \bf{85.69$\pm$1.55} & {89.54$\pm$0.49} 
        && 64.39$\pm$0.40 & {70.34$\pm$0.26}
        && 38.99$\pm$0.38\\
        ~~$\Delta$ (mean)
        && +2.98 & +2.97 & +0.42
        && +0.36 & +0.09
        && +0.21\\
        ~~$\Delta$ (max)
        && +3.65 & +5.19 & +0.70
        && +0.49 & +0.20
        && +0.92\\
        \midrule
        OpenMatch~\cite{saito2021openmatch}
        && 82.45$\pm$2.31 & 91.23$\pm$0.94 & 92.80$\pm$0.45
        && 70.23$\pm$0.30 & 74.56$\pm$0.46
        && 47.33$\pm$0.81\\
        ~~w/ WiseOpen-E $\dagger$
        && {83.69$\pm$1.59} & \bf{91.86$\pm$0.45} & 93.11$\pm$0.50 
        && 70.93$\pm$0.66 & {75.14$\pm$0.33}
        && {49.45$\pm$0.31}\\
        ~~w/ WiseOpen-E $\ddagger$
        && 83.35$\pm$1.95 & 91.47$\pm$0.53 & \bf{93.23$\pm$0.34} 
        && \bf{71.67$\pm$0.38} & 74.55$\pm$0.19
        && 49.14$\pm$0.33\\
        ~~w/ WiseOpen-L $\dagger$
        && 83.45$\pm$0.95 & {91.82$\pm$0.37} & {93.12$\pm$0.27} 
        && {71.23$\pm$0.59} & \bf{75.38$\pm$0.58}
        && \bf{49.75$\pm$0.69}\\
        ~~w/ WiseOpen-L $\ddagger$
        && \bf{84.69$\pm$0.76} & 91.34$\pm$0.69 & 92.93$\pm$0.06 
        && 71.12$\pm$0.31 & 75.09$\pm$0.43
        && 48.74$\pm$0.08\\
        ~~$\Delta$ (mean)
        && +1.34 & +0.39 & +0.30
        && +1.01 & +0.48
        && +1.94\\
        ~~$\Delta$ (max)
        && +2.24 & +0.63 & +0.43
        && +1.44 & +0.82
        && +2.42\\
        \midrule
        IOMatch~\cite{li2023iomatch}
        && 91.54$\pm$0.32 & 92.09$\pm$0.36 & 93.46$\pm$0.17
        && 69.83$\pm$0.59 & 73.87$\pm$0.25
        && 47.86$\pm$0.24\\
        ~~w/ WiseOpen-E $\dagger$
        && 91.78$\pm$0.17 & \bf{92.25$\pm$0.62} & \bf{93.59$\pm$0.07} 
        && \bf{70.49$\pm$0.28} & 74.24$\pm$0.41
        && 47.93$\pm$0.19 \\
        ~~w/ WiseOpen-E $\ddagger$
        && 91.77$\pm$0.08 & 92.16$\pm$0.18 & 93.36$\pm$0.16 
        && 69.97$\pm$0.55 & 74.12$\pm$0.12
        && 47.99$\pm$0.33 \\
        ~~w/ WiseOpen-L $\dagger$
        && \bf{91.90$\pm$0.16} & \bf{92.25$\pm$0.20} & 93.56$\pm$0.25 
        && 70.26$\pm$0.55 & \bf{74.36$\pm$0.45} 
        && 48.49$\pm$0.40 \\
        ~~w/ WiseOpen-L $\ddagger$
        && 91.16$\pm$0.29 & 92.02$\pm$0.20 & 93.35$\pm$0.08
        && 70.47$\pm$0.37 & 74.33$\pm$0.05 
        && \bf{49.18$\pm$0.40} \\
        ~~$\Delta$ (mean)
        && +0.11 & +0.08 & +0.00
        && +0.47 & +0.40
        && +0.54 \\
        ~~$\Delta$ (max)
        && +0.36 & +0.16 & +0.13
        && +0.67 & +0.49
        && +1.32 \\
        \bottomrule
    \end{tabular}
    
\end{table*}

\begin{table*}[!t]
    \centering
    \caption{Comparison of AUROC (in \%, mean $\pm$ standard deviation) for evaluating OOD detection performance. Higher is better.}
    \label{tb:main_results:roc}
    \begin{tabular}{l c ccc c cc c c}
        \toprule
         && \multicolumn{3}{c}{CIFAR-10} && \multicolumn{2}{c}{CIFAR-100} && \multicolumn{1}{c}{Tiny-ImageNet} \\
        \cmidrule{3-5} \cmidrule{7-8}  \cmidrule{10-10} 
        Algorithms && 50 labels & 100 labels & 400 labels && 50 labels & 100 labels  && 50 labels\\
        \midrule
        \noalign{\smallskip}
        Labeled Only
        && 56.15$\pm$1.81 & 59.58$\pm$2.25 & 69.95$\pm$0.60
        && 67.86$\pm$0.71 & 70.04$\pm$0.52
        && 61.49$\pm$0.38\\
        \midrule
        FixMatch~\cite{SohnBCZZRCKL20} 
        && 38.46$\pm$0.62 & 41.02$\pm$0.87 & \bf{48.49$\pm$1.41}
        && 60.19$\pm$0.30 & 63.14$\pm$0.23 
        && 58.65$\pm$0.82\\
        ~~w/ WiseOpen-E $\dagger$
        && \bf{39.26$\pm$0.97} & 41.20$\pm$1.56 &  47.53$\pm$0.44
        && \bf{60.63$\pm$1.07} & 62.57$\pm$0.12
        && \bf{59.41$\pm$0.70}\\
        ~~w/ WiseOpen-E $\ddagger$
        && 38.94$\pm$1.22 & \bf{42.26$\pm$1.10} & 48.30$\pm$1.03
        && 59.48$\pm$0.35 & \bf{63.28$\pm$0.52}
        && {59.23$\pm$0.69}\\
        ~~$\Delta$ (mean)
        && +0.64 & +0.70 & -0.58
        && -0.14 & -0.22
        && +0.67\\
        ~~$\Delta$ (max)
        && +0.79 & +1.23 & -0.19
        && +0.44 & +0.14
        && +0.76\\
        \midrule
        FreeMatch~\cite{Wang0HHFW0SSRS023} 
        && {45.94$\pm$1.84} & \bf{52.86$\pm$1.78} & \bf{64.67$\pm$1.12}
        && \bf{64.92$\pm$0.70} & {68.71$\pm$0.19} 
        && \bf{59.58$\pm$0.42}\\
        ~~w/ WiseOpen-E $\dagger$
        && \bf{47.38$\pm$2.03} & 51.65$\pm$2.27 & {62.75$\pm$1.88}
        && 64.27$\pm$0.38 & 67.38$\pm$0.29
        && 59.57$\pm$0.29\\
        ~~w/ WiseOpen-E $\ddagger$
        && 46.46$\pm$2.36 & 52.79$\pm$2.37 & 63.94$\pm$2.67
        && 64.22$\pm$0.89 & \bf{69.01$\pm$0.41}
        && 58.84$\pm$0.86\\
        ~~$\Delta$ (mean)
        && +0.98 & -0.64 & -1.33
        && -0.68 & -0.52
        && -0.38\\
        ~~$\Delta$ (max)
        && +1.44 & -0.07 & -0.73
        && -0.65 & +0.30
        && -0.01\\
        \midrule
        MTC~\cite{yu2020multi}
        && 77.77$\pm$1.10 & 80.36$\pm$2.13 & 87.02$\pm$0.91 
        && 65.40$\pm$0.60 & 64.58$\pm$0.26
        && 60.71$\pm$0.55\\
        ~~w/ WiseOpen-E $\dagger$
        && \bf{79.02$\pm$0.35} & \bf{82.02$\pm$1.98} & \bf{88.67$\pm$0.76}
        && \bf{66.78$\pm$1.79} & {65.09$\pm$1.25}
        && 61.08$\pm$0.62\\
        ~~w/ WiseOpen-E $\ddagger$
        && 78.10$\pm$0.36 & 79.08$\pm$1.74 & 86.59$\pm$2.45 
        && {65.97$\pm$0.91} & 64.41$\pm$1.25
        && 61.29$\pm$0.08\\
        ~~w/ WiseOpen-L $\dagger$
        && {78.85$\pm$0.63} & 81.64$\pm$1.36 & {88.14$\pm$0.88} 
        && 65.86$\pm$0.49 & 63.23$\pm$0.30
        && {61.44$\pm$0.28}\\
        ~~w/ WiseOpen-L $\ddagger$
        && 78.45$\pm$0.39 & {81.65$\pm$0.70} & 86.10$\pm$1.71 
        && 65.41$\pm$1.42 & \bf{65.57$\pm$1.14}
        && \bf{62.08$\pm$0.50}\\
        ~~$\Delta$ (mean)
        && +0.83 & +0.73 & +0.36
        && +0.60 & -0.00
        && +0.76\\
        ~~$\Delta$ (max)
        && +1.25 & +1.66 & +1.65
        && +1.38 & +0.99
        && +1.37\\
        \midrule
        OpenMatch~\cite{saito2021openmatch}
        && 58.70$\pm$8.71 & \bf{55.60$\pm$5.09} & 47.90$\pm$2.64
        && 73.82$\pm$0.16 & 74.58$\pm$0.59
        && 65.89$\pm$0.20\\
        ~~w/ WiseOpen-E $\dagger$
        && \bf{65.25$\pm$9.16} & 49.97$\pm$5.71 & \bf{53.10$\pm$4.32} 
        && {75.23$\pm$0.49} & {76.12$\pm$0.72}
        && 66.41$\pm$0.15\\
        ~~w/ WiseOpen-E $\ddagger$
        && {60.28$\pm$4.63} & {51.05$\pm$4.18} & {48.65$\pm$7.29} 
        && \bf{75.24$\pm$0.34} & 75.43$\pm$1.39
        && {66.84$\pm$0.35}\\
        ~~w/ WiseOpen-L $\dagger$
        && 55.33$\pm$4.99 & 49.70$\pm$4.45 & 44.14$\pm$3.55 
        && 74.59$\pm$0.40 & \bf{76.26$\pm$1.02}
        && 66.71$\pm$0.57\\
        ~~w/ WiseOpen-L $\ddagger$
        && 58.40$\pm$4.77 & 47.31$\pm$3.68 & 43.95$\pm$1.74
        && 74.88$\pm$0.72 & 74.92$\pm$1.63
        && \bf{67.33$\pm$0.21}\\
        ~~$\Delta$ (mean)
        && +1.11 & -6.09 & -0.44
        && +1.17 & +1.10
        && +0.93\\
        ~~$\Delta$ (max)
        && +6.55 & -4.55 & +5.20
        && +1.42 & +1.68
        && +1.44\\
        \midrule
        IOMatch~\cite{li2023iomatch}
        && \bf{44.54$\pm$0.29} & {48.02$\pm$0.87} & 61.80$\pm$2.33
        && \bf{67.44$\pm$0.88} & {69.49$\pm$0.32}
        && 62.73$\pm$0.28\\
        ~~w/ WiseOpen-E $\dagger$
        && 42.51$\pm$2.22 & 48.34$\pm$0.96 & 61.49$\pm$2.17 
        && 65.99$\pm$0.14 & 69.36$\pm$0.48 
        && \bf{62.96$\pm$0.56} \\
        ~~w/ WiseOpen-E $\ddagger$
        && 43.10$\pm$1.12 & 48.23$\pm$0.45 & \bf{63.45$\pm$1.04} 
        && 66.74$\pm$0.50 & \bf{69.67$\pm$0.10}
        && 62.43$\pm$0.22 \\
        ~~w/ WiseOpen-L $\dagger$
        && 44.23$\pm$1.74 & \bf{48.45$\pm$1.05} & 60.58$\pm$1.18 
        && 67.04$\pm$0.43 & 69.43$\pm$0.53 
        && 62.66$\pm$0.60 \\
        ~~w/ WiseOpen-L $\ddagger$
        && 42.67$\pm$0.54 & 47.60$\pm$1.23 & 60.90$\pm$1.35 
        && 66.83$\pm$0.45 & {69.63$\pm$0.32} 
        && 62.20$\pm$0.64 \\
        ~~$\Delta$ (mean)
        && -1.41 & +0.13 & -0.20
        && -0.79 & +0.03
        && -0.17\\
        ~~$\Delta$ (max)
        && -0.31 & +0.43 & +1.65
        && -0.40 & +0.18
        && +0.23\\
        \bottomrule
    \end{tabular}  
\end{table*}

To present the way we construct the OSSL scenario more specifically, we take the OSSL scenario constructing procedure on CIFAR-10 as an example. In our experiments, we take the first $6$ classes of CIFAR-10 as seen classes, namely airplane, automobile, bird, cat, deer, and dog,  while the remaining $4$ classes consisting of frog, horse, ship, and truck are taken as unseen classes. We randomly sample the images in the training set to construct a labeled set, an unlabeled set, and a validation set. Considering the setting of CIFAR-10 with 100 labels per class, we will get a labeled training set of 600 images with annotation in total which come from the seen classes, an unlabeled training set of 49,100 images where 29,100 images are from the seen classes while the rest 20,000 images are from the unseen classes, a validating set of 300 images from the seen classes, and a testing set of 10,000 images from the seen and unseen classes. 

\textbf{Baselines and Evaluation.} There are three categories of the baseline methods: (1) Labeled Only method, which only trains the model with labeled data using cross-entropy loss; (2) traditional SSL methods, including FixMatch~\cite{SohnBCZZRCKL20} and FreeMatch~\cite{Wang0HHFW0SSRS023}; (3) OSSL methods, including MTC~\cite{yu2020multi}, OpenMacth~\cite{saito2021openmatch}, and IOMatch~\cite{li2023iomatch}. As for the evaluation of ID classification, we employ the top-1 accuracy over the testing instances from seen classes. Moreover, to evaluate OOD detection performance, the AUROC~\cite{DavisG06} over the whole testing set is adopted following~\cite{hendrycksbaseline, saito2021openmatch}.

\textbf{Implementation Details.} For fairness, followed by~\cite{saito2021openmatch, yu2020multi, li2023iomatch}, Wide ResNet-28-2~\cite{ZagoruykoK16} is employed as the backbone representation extractor for all methods. For MTC, we apply their official implementation with Pytorch. For other methods, we utilize~\url{https://github.com/kekmodel/FixMatch-pytorch} as their pipeline to conduct the experiments. The total number of training epochs for CIFAR-10/100 and Tiny-ImageNet are 512 and 256 respectively, and each epoch consists of 1024 iterations. For all experiments, the batch sizes for labeled data and unlabeled data are 64 and 128 respectively. We employ the SGD with a nesterov momentum of 0.9 as the optimizer, and the learning rate is initialized as 0.03 and decays in a cosine annealing manner. 
The hyper-parameter $k$ utilized in Top-k thresholds for different scenarios is summarised in Table~\ref{tb:topk}. All experiments can be performed on a single NVIDIA GeForce RTX 4090.

\subsection{Main Results}
\label{results}

As shown in Figure~\ref{fig:wiseopen}, we first conduct the experiments on top of OpenMatch on CIFRA-100 with 100 labels per seen class to validate the effectiveness of WiseOpen. We can observe that WiseOpen can make respectable performance improvements to the original OpenMatch model. However, as shown in Table~\ref{tb:runing_time}, WiseOpen is relatively time-consuming on account of the frequent GV-SM that requires the computation of gradient variance for each unlabeled instance in every epoch. Therefore, to extensively validate our proposed method, together with the consideration of the computational cost, we compare all SSL and OSSL baselines with WiseOpen's variants on top of their original algorithm. Note that FixMatch and FreeMatch can actually be seen as two variants of WiseOpen-L with their specific techniques as we mentioned in Section~\ref{sec:relation}. Thus, for the SSL methods, we omit the WiseOpen-L experiments and only conduct WiseOpen-E experiments. The results are reported in Table~\ref{tb:main_results:acc} and Table~\ref{tb:main_results:roc}, where the number of labels per seen class is provided for each column. In all settings, the first 60\% of classes are taken as seen classes. From these results, we can obtain the following observations: 
\begin{enumerate}
    \item [(1)] Focusing on the ID classification performance among different methods, the Labeled Only method usually has the poorest performance, because of its no access to the open-set data. Being beneficial from the friendly data in the open-set data, traditional SSL methods and OSSL methods can achieve better performance in most cases, while our proposed frameworks can usually further enhance their performance since we exclude some unfriendly open-set data in the training procedure.
    \item [(2)] Comparing the ID classification performance among our proposed WiseOpen and its two variants, we can observe that WiseOpen can make more sound improvements, particularly when the threshold is an adaptive threshold like the Otsu threshold. Meanwhile, WiseOpen-E and WiseOpen-L are also competent in improving the performance of the models in most scenarios with more acceptable extra training costs.
    \item [(3)] By simultaneously considering the performance of ID classification and OOD detection, it appears that there is no necessary correlation between these two abilities. The algorithms with the best ID classification performance do not consistently exhibit excellent OOD detection performance and sometimes even perform poorly in distinguishing the ID instances from OOD instances. This phenomenon implicitly shows that the key point of handling the open-set data to enhance the ID classification performance lies in selecting and exploiting the friendly open-set data rather than simply detecting and discarding all OOD instances.
\end{enumerate}

In summary, the main experiment results convincingly demonstrate the effectiveness of our proposed frameworks and correspond with our core idea that we should wisely leverage the open-set data to obtain valuable from the friendly open-set data while preventing being contaminated by the unfriendly ones.

\begin{figure}[!t]
    \centering
    \subfigure[WiseOpen-E $\ddagger$ on top of OpenMatch]{
        \centering
        \includegraphics[width=0.95\textwidth]{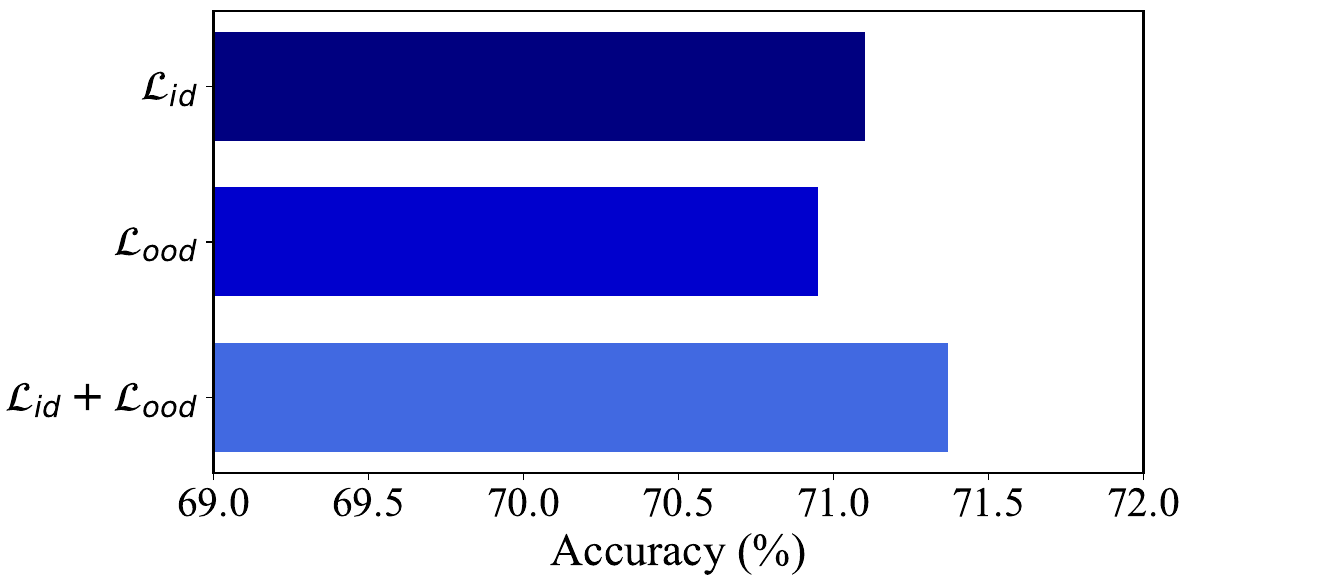}
    }
    \vspace{1mm}
    \subfigure[WiseOpen-E $\ddagger$ on top of IOMatch]{
        \centering
        \includegraphics[width=0.95\textwidth]{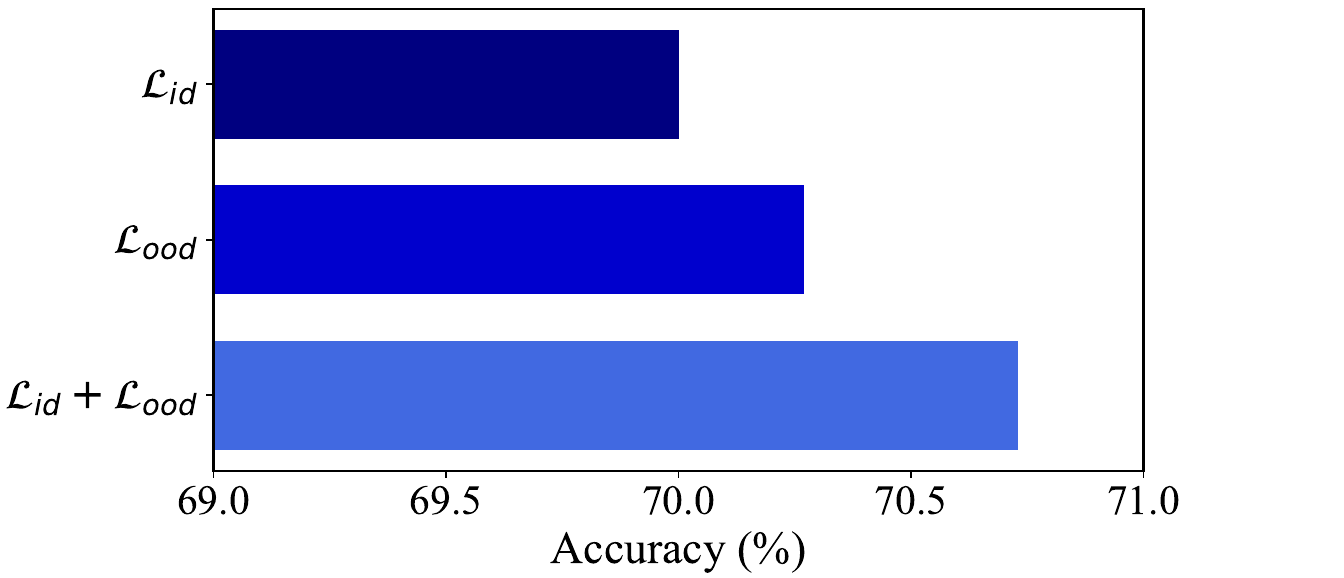}
    }
    \caption{ID Classification performance of WiseOpen-E $\ddagger$ with different losses used in GV-SM on top of OpenMatch and IOMatch. Models are trained with CIFAR-100 at 50 labels.}
    \label{fig:ab_loss}
\end{figure}

\subsection{Ablation Study}
\label{ablation}
\textbf{Comparison of GV-SM by Different Losses.} OSSL methods with OOD detection modules typically incorporate two components in their unsupervised loss to optimize the modeling in the ID domain and the open-set domain respectively. Here we name the two components as $\mathcal{L}_{id}$ and $\mathcal{L}_{ood}$ respectively. For example, in OpenMatch, $\mathcal{L}_{id}$ and $\mathcal{L}_{ood}$ will be $\mathcal{L}_{fm}$ in Eq.\ref{eqn:loss:fm}, and the sum of $\mathcal{L}_{em}$ in Eq.\ref{eqn:loss:em} and $\mathcal{L}_{oc}$ in Eq.\ref{eqn:loss:oc} respectively. For IOMatch, $\mathcal{L}_{id}$ and $\mathcal{L}_{ood}$ will be the $\mathcal{L}_{ui}$ and $\mathcal{L}_{op}$, which are the cross-entropy loss for $K$-classification and $(K+1)$-classification by considering all unseen classes as a single novel class respectively. \jn{We perform the experiments of WiseOpen-E $\ddagger$ on top of OpenMatch and IOMatch with different losses used in GV-SM, using CIFAR-100 at 50 labels.} As shown in
Figure~\ref{fig:ab_loss}, we can observe that utilizing the whole unsupervised loss can promote the model to better performance compared with using single $\mathcal{L}_{id}$ or $\mathcal{L}_{ood}$, which is corresponding to our theoretical analysis.

\textbf{Sensitivity Analysis on Mismatching Ratio.} To further demonstrate the robustness of our proposed frameworks, we conduct experiments \jn{on CIFAR-100 at 50 labels per class} with varying class-mismatching ratios. Specifically, we evaluate the ID classification performance of WiseOpen-E and WiseOpen-L on top of OpenMatch and IOMacth along with the vanilla OpenMatch and IOMacth with varying values of $K\in\{60, 50, 40, 30\}$, \emph{i.e.}, the mismatching ratio varies from 0.4 to 0.7. The experimental results are summarised in Table~\ref{tb:varying_ratio}. We can observe that our proposed frameworks can enhance the ID classification capability in most cases, which demonstrates that our proposed frameworks are robust across different class-mismatching ratios.

\begin{table}[!t]
    \caption{ID classification accuracy (in \%) of WiseOpen-E and WiseOpen-L on top of OpenMatch and IOMatch in varying mismatching scenarios of CIFAR100 with 50 labels.}
    \label{tb:varying_ratio}   
    \centering
    \setlength{\tabcolsep}{8pt}
    \begin{tabular}{lcccc}
        \toprule
        Mismatching ratios & 0.4 & 0.5 & 0.6 & 0.7\\
        \midrule
        \noalign{\smallskip}
        OpenMatch
        & 70.53 & 73.18 & 76.18 & 79.20 \\
        ~~w/ WiseOpen-E $\dagger$ 
        & 71.85 & 73.68 & 77.40 & 78.93\\
        ~~w/ WiseOpen-E $\ddagger$
        & 71.37 & \bf{74.06} & 76.17 & 79.20\\
        ~~w/ WiseOpen-L $\dagger$
        & \bf{71.88} & {73.76} & \bf{77.70} & \bf{79.43}\\
        ~~w/ WiseOpen-L $\ddagger$
        & 71.53 & 73.14 & 76.70 & 78.67 \\
        ~~$\Delta$ (mean)
        & +1.13 & +0.48 & +0.81 & -0.14 \\
        ~~$\Delta$ (max)
        & +1.35 & +0.88 & +1.52 & +0.23 \\
        \midrule
        IOMatch
        & 70.53 & 72.74 & 74.98 & 77.63 \\
        ~~w/ WiseOpen-E $\dagger$
        & 70.88 & \bf{73.20} & \bf{75.52} & 76.90\\
        ~~w/ WiseOpen-E $\ddagger$
        & 70.73 & 73.04 & 75.35 & 77.20 \\
        ~~w/ WiseOpen-L $\dagger$
        & \bf{71.02} & 72.90 & 75.32 & \bf{78.27} \\
        ~~w/ WiseOpen-L $\ddagger$
        & 70.93 & 73.02 & 75.07 & 78.13\\
        ~~$\Delta$ (mean)
        & +0.36& +0.30 & +0.34 & -0.00\\
        ~~$\Delta$ (max)
        & +0.49 & +0.46 & +0.54 & +0.64\\
        \bottomrule
    \end{tabular}
    \vspace{2mm}
\end{table}

\begin{table*}[!t]
    \begin{center}
    \caption{ID classification accuracy (in \%) employing Adam optimizer and RMSProp optimizer. 50 labels per seen class are utilized in training models.}
    \label{tb:opt}
    \resizebox{0.8\columnwidth}{!}{
    \begin{tabular}{l c cc c cc c cc}
        \toprule
        && \multicolumn{2}{c}{CIFAR-10} && \multicolumn{2}{c}{CIFAR-100} && \multicolumn{2}{c}{Tiny-ImageNet} \\
    \cmidrule{3-4} \cmidrule{6-7}  \cmidrule{9-10} 
    Algorithms && Adam & RMSProp && Adam & RMSProp  && Adam & RMSProp\\
        \midrule
        \noalign{\smallskip}
        OpenMatch
        && 78.78 &  78.57
        && 70.07 &  68.83
        && 48.10 &  47.42\\
        ~~w/ WiseOpen-E $\dagger$ 
        && 73.10 &  76.88
        && 69.77 &  69.52
        && 47.98 &  \bf{47.98}\\
        ~~w/ WiseOpen-E $\ddagger$
        && 74.00 & 76.20 
        && 70.28 & 68.45
        && 47.77 & 46.90\\
        ~~w/ WiseOpen-L $\dagger$
        && \bf{81.67} &  78.93
        && 71.12 &  68.40
        && 48.18 &  47.13\\
        ~~w/ WiseOpen-L $\ddagger$
        && 81.07 &  \bf{82.40}
        && \bf{71.47} &  \bf{69.97}
        && \bf{48.35} &  47.57\\
        ~~$\Delta$ (mean)
        && -1.32 &  +0.04
        && +0.59 &  +0.25
        && -0.03 & -0.02 \\
        ~~$\Delta$ (max)
        && +2.88 &  +3.83
        && +1.40 & +1.13
        && +0.25 & +0.57 \\
        \midrule
        IOMatch
        && 90.27 &  91.78
        && 70.60 &  70.32
        && 45.35 &  45.28\\
        ~~w/ WiseOpen-E $\dagger$ 
        && \bf{91.37} &  91.18
        && 70.75 &  \bf{70.73}
        && 46.33 &  45.03\\
        ~~w/ WiseOpen-E $\ddagger$
        && 90.55 &  91.83
        && \bf{70.85} &  70.53
        && 46.13 &  44.63\\
        ~~w/ WiseOpen-L $\dagger$
        && 90.00 &  \bf{92.32}
        && 70.83 &  70.62
        && 46.00 &  \bf{45.30}\\
        ~~w/ WiseOpen-L $\ddagger$
        && 89.75 &  91.05
        && 70.68 &  70.70
        && \bf{46.40} &  45.23\\
        ~~$\Delta$ (mean)
        && +0.15 &  -0.19
        && +0.18 &  +0.33
        && +0.87 &  -0.23\\
        ~~$\Delta$ (max)
        && +1.10 &  +0.53
        && +0.25 &  +0.42
        && +1.05 &  +0.02\\
        \bottomrule
    \end{tabular}
    }
    \end{center}
\end{table*}

\jn{
\textbf{Effects of Various Optimizers.} As summarized in Table~\ref{tb:opt}, we provide the ID classification performance of WiseOpen-E and WiseOpen-L on top of OpenMatch and IOMacth along with their vanilla versions, using the Adam optimizer and RMSProp optimizer. The initial learning rate is set as 0.0003 across all experiments and other hyper-parameters remain the same as reported in Section~\ref{expe:settings}. 50 labels per class are utilized in all datasets. As we can observe, our proposed methods successfully make improvements in most benchmarks but the performance of WiseOpen-E is not as stable as when using the SGD optimizer. This phenomenon is reasonable since our proposed GV-SM is built upon the theoretical analysis based on the stochastic gradient used in SGD.
}

\begin{table*}[!htbp]
    \begin{center}
    \caption{Evaluation of OOD detection on OOD data unseen in the training set (AUROC in \%). Models are trained on Tiny-ImageNet with 50 labeled data per class.}
    \label{tb:unseen_ood}
    \resizebox{1.0\columnwidth}{!}{
        \begin{tabular}{lccccccc}
            \toprule
             & \multicolumn{6}{c}{Unseen OOD Datasets} \\
            \cmidrule(lr){2-7}
            Algorithms & LSUN  & DTD & CUB & Flowers & Caltech & Dogs & MEAN \\
            \midrule
            MTC & 37.51$\pm$1.40 & 35.25$\pm$2.60 & 47.91$\pm$3.48 & 52.28$\pm$2.79 & 47.49$\pm$2.93 & 40.24$\pm$4.99 & 43.45$\pm$6.94  \\
            ~~w/ WiseOpen-E $\dagger$ & 39.39$\pm$7.39 & 37.73$\pm$2.75 & 48.70$\pm$0.72 & 55.24$\pm$3.04 & 51.69$\pm$0.79 & 44.01$\pm$3.36 & 46.13$\pm$7.36  \\
            ~~w/ WiseOpen-E $\ddagger$ & 41.10$\pm$11.44 & 37.17$\pm$1.31 & 48.38$\pm$4.21 & {49.60$\pm$3.29} & 50.00$\pm$3.49 & {36.76$\pm$1.44} & 43.84$\pm$7.85  \\
            ~~w/ WiseOpen-L $\dagger$ & {34.82$\pm$2.45} & 35.93$\pm$0.86 & 50.06$\pm$3.25 & 54.41$\pm$6.50 & 50.87$\pm$2.83 & 43.24$\pm$1.65 & 44.89$\pm$8.25  \\
            ~~w/ WiseOpen-L $\ddagger$ & \bf{45.43$\pm$15.31} & \bf{47.81$\pm$2.07} & \bf{58.03$\pm$4.75} & \bf{60.51$\pm$2.96} & \bf{57.26$\pm$4.41} & \bf{48.85$\pm$1.24} & \bf{52.98$\pm$9.06} \\
            ~~$\Delta$ (mean) & +2.68 & +4.41 & +3.38 & +2.66 & +4.96 & +2.97& +3.51\\
            ~~$\Delta$ (max) & +7.92 & +12.56 & +10.11 & +8.23 & +9.77  & +8.61 & +9.53\\
            \midrule
            OpenMatch & 53.06$\pm$2.72 & 46.84$\pm$0.20 & 57.04$\pm$0.15 & 55.88$\pm$1.41 & 60.00$\pm$0.92 & \bf{61.13$\pm$0.67} & 55.66$\pm$4.93  \\
            ~~w/ WiseOpen-E $\dagger$  & 54.38$\pm$1.24 & 47.68$\pm$1.38 & 58.45$\pm$1.14 & {55.72$\pm$2.30} & 61.84$\pm$1.06 & {59.77$\pm$1.34} & 56.31$\pm$4.81  \\
            ~~w/ WiseOpen-E $\ddagger$ & \bf{56.41$\pm$0.98} & 49.20$\pm$2.01 & 59.57$\pm$1.62 & 57.02$\pm$2.02 & 62.43$\pm$0.67 & {59.61$\pm$1.31} & 57.37$\pm$4.42  \\
            ~~w/ WiseOpen-L $\dagger$ & 56.22$\pm$0.78 & \bf{49.60$\pm$0.88} & 59.39$\pm$0.20 & 59.62$\pm$1.17 & \bf{62.97$\pm$0.61} & {59.81$\pm$1.46} & \bf{57.93$\pm$4.31}   \\
            ~~w/ WiseOpen-L $\ddagger$ & 56.01$\pm$2.64 & 49.54$\pm$3.20 & \bf{59.97$\pm$0.14} & \bf{59.81$\pm$1.91} & 62.10$\pm$1.23 & {58.89$\pm$1.53} & 57.72$\pm$4.56  \\
            ~~$\Delta$ (mean) & +2.70 & +2.16 & +2.30 & +2.16 & +2.33 & -1.61 & +1.68 \\
            ~~$\Delta$ (max) & +3.35 & +2.76 & +2.93 & +3.93 & +2.97 & -1.32 & +2.28 \\
            \midrule
            IOMatch & 63.88$\pm$1.76 & 56.53$\pm$2.11 & 63.10$\pm$2.07 & 64.52$\pm$4.22 & 63.71$\pm$1.15 & \bf{63.71$\pm$0.53} & 62.57$\pm$3.56  \\
            ~~w/ WiseOpen-E $\dagger$ & \bf{67.28$\pm$0.59} & 57.46$\pm$1.65 & 65.14$\pm$1.54 & 67.37$\pm$0.38 & {62.87$\pm$0.34} & {62.77$\pm$1.48} & 63.81$\pm$3.57 \\
            ~~w/ WiseOpen-E $\ddagger$ & 65.47$\pm$1.67 & \bf{59.75$\pm$0.60} & 64.23$\pm$1.03 & 66.29$\pm$3.21 & {63.31$\pm$0.64} & {62.98$\pm$1.39} & 63.67$\pm$2.68  \\
            ~~w/ WiseOpen-L $\dagger$ & 66.16$\pm$3.31 & 57.07$\pm$2.41 & \bf{66.56$\pm$0.84} & 67.04$\pm$2.40 & \bf{63.73$\pm$1.03} & {63.41$\pm$0.98} & \bf{63.99$\pm$3.96}   \\
            ~~w/ WiseOpen-L $\ddagger$ & 64.93$\pm$2.04 & 57.93$\pm$2.35 & 64.18$\pm$1.56 & \bf{68.85$\pm$0.69} & {63.65$\pm$0.42} & {62.93$\pm$1.09} & 63.75$\pm$3.56  \\
            ~~$\Delta$ (mean) & +2.08 & +1.52 & +1.93 & +2.86& -0.32 & -0.69 & +1.23\\
            ~~$\Delta$ (max) & +3.40 & +3.22 & +3.47 & +4.33& +0.02 & -0.30 & +1.42\\
            \bottomrule
        \end{tabular}
    }

    \end{center}
\end{table*}

\begin{table*}[!t]
    \begin{center}
    \caption{Comparison of ID classification accuracy (in \%) among IOMatch (reduced), the proposed WiseOpen variants on top of IOMatch (reduced), and full IOMatch. 50 labels per seen class are utilized in training models.}
    \label{tb:nim}
    \begin{tabular}{l c c c}
        \toprule
        Datasets & CIFAR-10 & CIFAR-100 & Tiny-ImageNet \\
        \midrule
        \noalign{\smallskip}
        IOMatch
        & 91.90 & 70.53 & 47.68\\
        \midrule
        IOMatch (reduced)
        & 91.47 & 69.97 & 46.75 \\
        ~~w/ WiseOpen-E $\dagger$ 
        & 91.40 & 70.85 & 47.13 \\
        ~~w/ WiseOpen-E $\ddagger$
        & 91.18 & 70.67 & 47.00\\
        ~~w/ WiseOpen-L $\dagger$
        & 91.55 & 70.53 & 46.83\\
        ~~w/ WiseOpen-L $\ddagger$
        & 91.40 & 70.47 & 47.37\\ 
        \bottomrule
    \end{tabular}
    \end{center}
\end{table*}

\subsection{Expansion on OOD Detection}
\label{unseen_ood}

Except for solving the ID classification problem, it's worth noting that models trained using most OSSL algorithms typically also develop the capability of OOD detection. In this subsection, we aim to explore the impact of our proposed frameworks on the models' capacity to detect various OOD instances. Specifically, we evaluate the OOD detection performance of our proposed frameworks on the OOD datasets that are not encountered during the training procedure. Following~\cite{tack2020csi}, we employ six benchmarks as OOD datasets unseen in training, namely, LSUN~\cite{YuZSSX15}, DTD~\cite{CimpoiMKMV14}, CUB-200~\cite{wah2011caltech}, Flowers~\cite{NilsbackZ06}, Caltech~\cite{griffin2007caltech}, and Dogs~\cite{khosla2011novel}, for models trained on Tiny-ImageNet with 50 labels per seen class. AUROC is employed to evaluate the detection performance. As summarized in Table~\ref{tb:unseen_ood}, we can observe that our proposed frameworks can promote the OOD detection performance of the majority of the unseen OOD instances. For instance, in terms of the mean OOD detection performance across all unseen datasets, our frameworks demonstrate an average improvement of 3.51\% and a maximum improvement of 9.53\% for MTC. This suggests that our proposal is sufficiently safe and will not compromise the potential OOD detection capability of the original model; in fact, it may even enhance it.

\begin{figure*}[!htbp]
    \centering
    \subfigure[Confusion matrices of WiseOpen-E $\ddagger$ on top of OpenMatch BEFORE GV-SM in epoch 50, 250, 450, accordingly.]{
        \centering
        \includegraphics[width=0.33\textwidth]{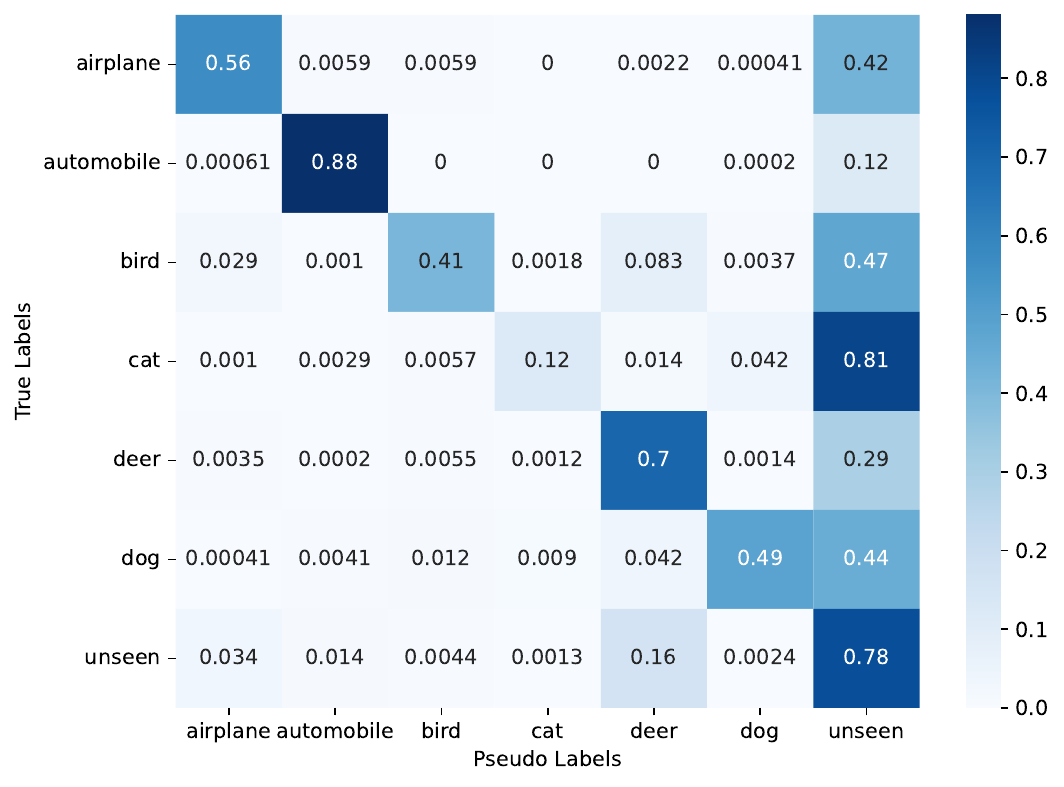}
        \includegraphics[width=0.33\textwidth]{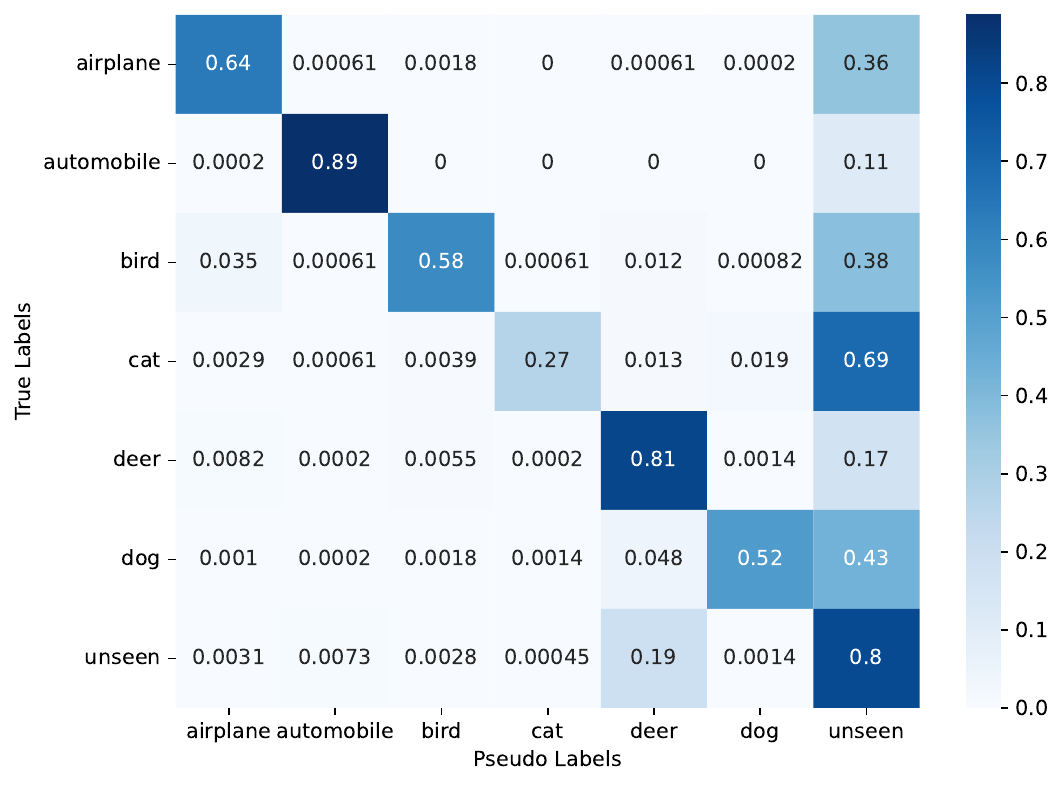}
        \includegraphics[width=0.33\textwidth]{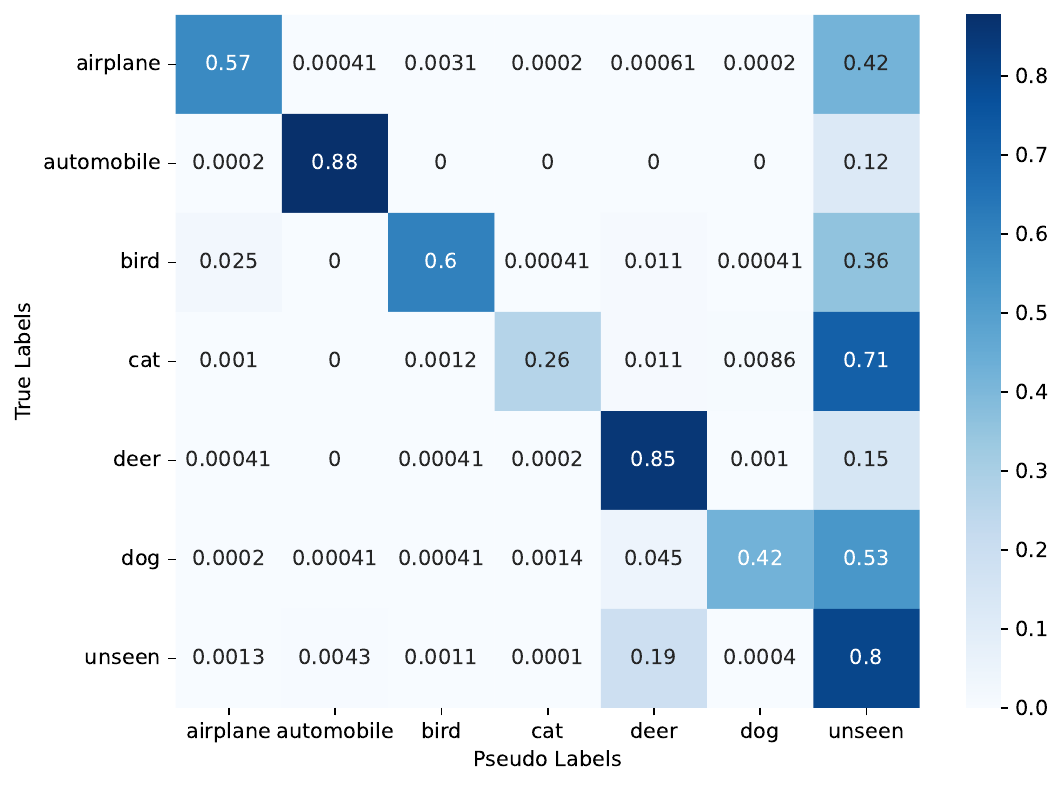}
    }
    \subfigure[Confusion matrices of WiseOpen-E $\ddagger$ on top of OpenMatch AFTER GV-SM in epoch 50, 250, 450, accordingly.]{
        \centering
        \includegraphics[width=0.33\textwidth]{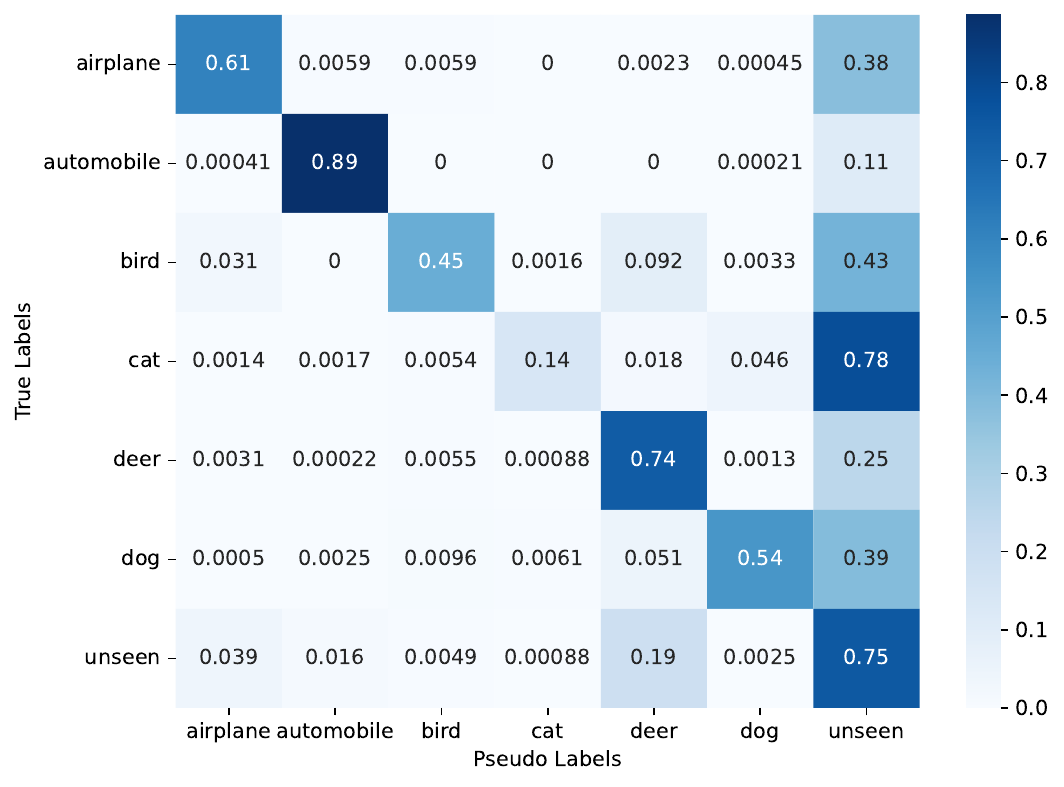}
        \includegraphics[width=0.33\textwidth]{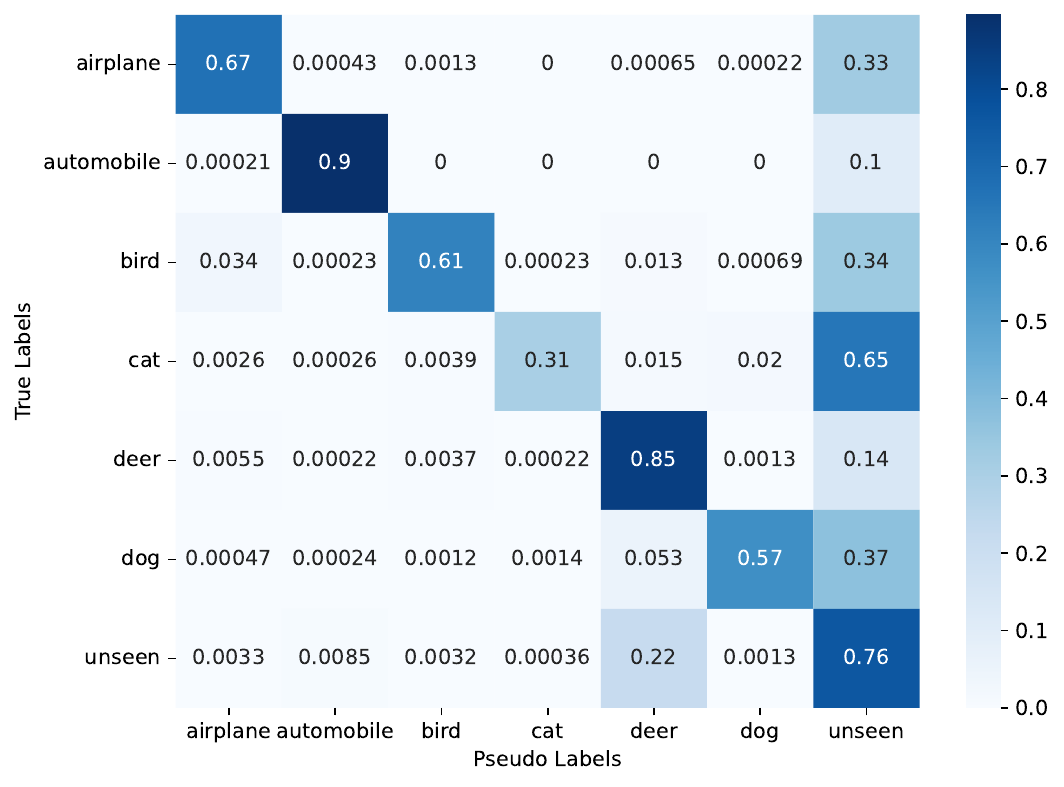}
        \includegraphics[width=0.33\textwidth]{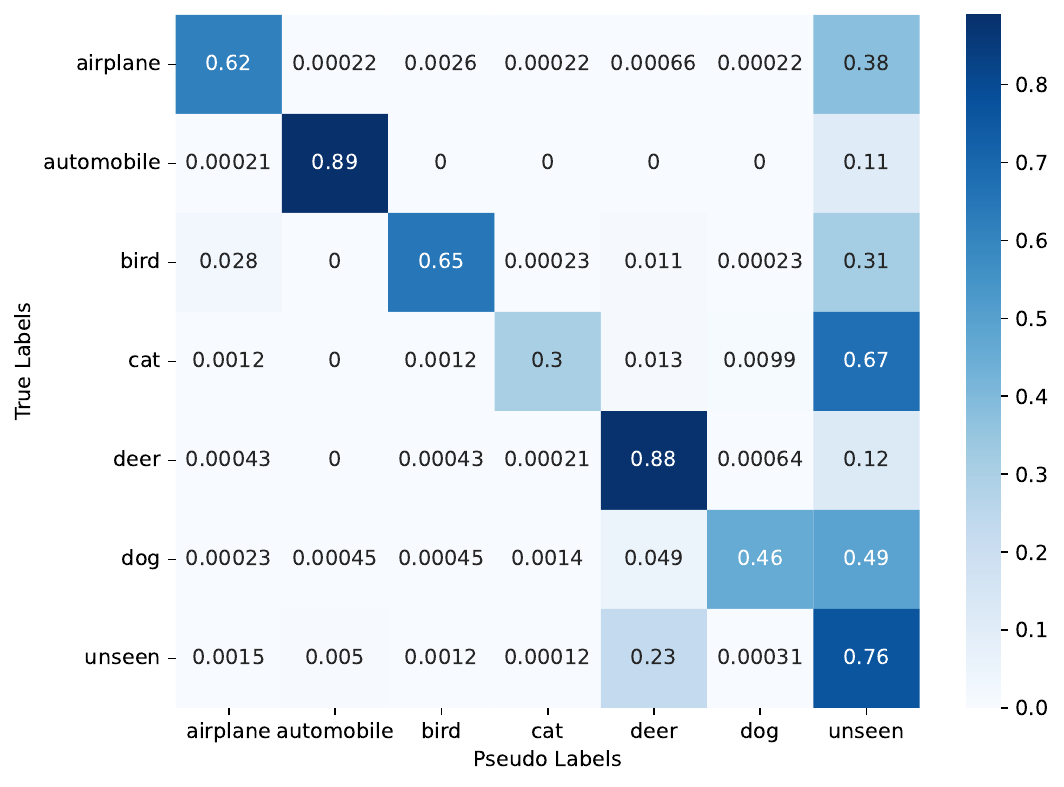}
    }
    \caption{Visualization of the confusion matrices on the unlabeled training set of CIFAR-10. }
    \label{fig:confusion}
\end{figure*}
\vspace{-1mm}
\jn{\subsection{Further Analysis}}
\vspace{-1mm}
\jn{\textbf{Comparison between Existing Selection Mechanisms.} Existing OSSL methods typically apply some OOD detection techniques as selection mechanisms tailored for one specific loss. Previous experimental results in Table~\ref{tb:main_results:acc} have successfully demonstrated that our proposed selection mechanism is orthogonal and complementary to prior selection mechanisms that can improve previous algorithms' performance after effortlessly embedding our proposal. In consideration of the completeness, we further perform the experiments of IOMatch (reduced), which omits the OOD data exclusion process while calculating the ID classification loss $\mathcal{L}_{ui}$. As summarized in Table~\ref{tb:nim}, we can observe that our proposed selection mechanism can improve the ID classification performance of IOMatch (reduced) in most cases. However, IOMatch (reduced) plus our selection mechanisms do not always outperform the full IOMatch. The reason may lie in that the OOD data exclusion is specifically tailored for the unlabeled inlier loss and the omission of such exclusion can potentially introduce biases to ID classification by misclassifying OOD data to seen classes. Our selection mechanism can alleviate the influence and improve the model's performance by selectively leveraging friendly data. However, in some datasets, the negative impacts of discarding OOD detection may be greater than the positive effect brought by our selection mechanism. Therefore, in some instances, the OOD data exclusion may appear to be more effective, while in others, our proposed selection mechanism may lead to more significant improvements. Overall, these empirical results show that although not tailored for the specific loss, our proposal is effective and robust, and can further enhance the model's capability which already utilizes the tailored selection in the training procedure.}

\textbf{Analysis of GV-SM.} To further analyse the effectiveness of our proposal,
Figure~\ref{fig:confusion} visualizes the confusion matrix of the unlabeled training set before and after applying GV-SM. We can observe that after applying GV-SM, the model can utilize the open-set data set with higher pseudo-labeling accuracy on the seen classes. Meanwhile, the mismatching from unseen classes to the seen class \emph{deer} slightly increases. Upon further investigation, we have discovered that among the unseen classes frog, horse, ship, and truck, the mismatch from horse to deer contributes to the overwhelming majority of mismatches from unseen classes to deer with ratios reaching 98.41\%, 99.27\%, and 99.81\% for epochs 50, 250, and 450, respectively. Intuitively, this type of mismatch is unlikely to negatively impact the distinction between deer and other seen classes, and it may even enhance this distinction by facilitating the learning of similar representations between horses and deer, which are distinct from other seen classes.

\vspace{3mm}

\section{Conclusion}
\label{conc}

In this paper, we tackled the realistic OSSL scenario, where models can suffer from performance degradation if recklessly utilize all open-set data which contain instances from unseen classes. Motivated by theoretical insights, we highlight the significance of selectively leveraging open-set data and propose a novel OSSL framework called WiseOpen which wisely leverages the open-set data by using GV-SM. GV-SM enables the model to exclude potentially unfriendly open-set training data with large gradient variance, thereby helping to maintain the purity of the learned knowledge. Furthermore, we also proposed two variants of WiseOpen dubbed WiseOpen-E and WiseOpen-L to mitigate the huge computing cost issue. Sufficient empirical results have demonstrated the effectiveness of our proposal. 

\section*{Acknowledgments}
This work was supported by National Key RD Program of China (2022YFF0712100), NSFC (62276131, 61921006, 08120002), the Fundamental Research Funds for the Central Universities (No.30922010317), the Fundamental Research Funds for the Central University of China (DUT No. 82232031), and Collaborative Innovation Center of Novel Software Technology and Industrialization.

\bibliographystyle{IEEEtran}
\bibliography{IEEEabrv, ref}

\begin{thebibliography}{10}
\providecommand{\url}[1]{#1}
\csname url@samestyle\endcsname
\providecommand{\newblock}{\relax}
\providecommand{\bibinfo}[2]{#2}
\providecommand{\BIBentrySTDinterwordspacing}{\spaceskip=0pt\relax}
\providecommand{\BIBentryALTinterwordstretchfactor}{4}
\providecommand{\BIBentryALTinterwordspacing}{\spaceskip=\fontdimen2\font plus
\BIBentryALTinterwordstretchfactor\fontdimen3\font minus \fontdimen4\font\relax}
\providecommand{\BIBforeignlanguage}[2]{{%
\expandafter\ifx\csname l@#1\endcsname\relax
\typeout{** WARNING: IEEEtran.bst: No hyphenation pattern has been}%
\typeout{** loaded for the language `#1'. Using the pattern for}%
\typeout{** the default language instead.}%
\else
\language=\csname l@#1\endcsname
\fi
#2}}
\providecommand{\BIBdecl}{\relax}
\BIBdecl

\bibitem{CSZ2006}
O.~Chapelle, B.~Sch{\"{o}}lkopf, and A.~Zien, Eds., \emph{Semi-Supervised Learning}.\hskip 1em plus 0.5em minus 0.4em\relax The {MIT} Press, 2006.

\bibitem{zhu2009introduction}
X.~Zhu and A.~B. Goldberg, ``Introduction to semi-supervised learning,'' \emph{Synth. Lect. Artif. Intell. Mach. Learn.}, vol.~3, no.~1, pp. 1--130, 2009.

\bibitem{li2019towards}
Y.-F. Li, L.-Z. Guo, and Z.-H. Zhou, ``Towards safe weakly supervised learning,'' \emph{IEEE Trans. Pattern Anal. Mach. Intell.}, vol.~43, no.~1, pp. 334--346, Jun. 2019.

\bibitem{zhou2018brief}
Z.-H. Zhou, ``A brief introduction to weakly supervised learning,'' \emph{Nat. Sci. Rev.}, vol.~5, no.~1, pp. 44--53, 2018.

\bibitem{yang2021semi1}
Y.~Yang, Z.-Y. Fu, D.-C. Zhan, Z.-B. Liu, and Y.~Jiang, ``Semi-supervised multi-modal multi-instance multi-label deep network with optimal transport,'' \emph{IEEE Trans. Knowl. Data Eng.}, vol.~33, no.~2, pp. 696--709, Feb. 2021.

\bibitem{yang2023cost}
Y.~Yang, D.-W. Zhou, D.-C. Zhan, H.~Xiong, Y.~Jiang, and J.~Yang, ``Cost-effective incremental deep model: Matching model capacity with the least sampling,'' \emph{IEEE Trans. Knowl. Data Eng.}, vol.~35, no.~4, pp. 3575--3588, Apr. 2023.

\bibitem{LaineA17}
S.~Laine and T.~Aila, ``Temporal ensembling for semi-supervised learning,'' in \emph{Proc. Int. Conf. Learn. Representations}, 2017.

\bibitem{MiyatoMKI19}
T.~Miyato, S.~Maeda, M.~Koyama, and S.~Ishii, ``Virtual adversarial training: {A} regularization method for supervised and semi-supervised learning,'' \emph{{IEEE} Trans. Pattern Anal. Mach. Intell.}, vol.~41, no.~8, pp. 1979--1993, Aug. 2019.

\bibitem{TarvainenV17}
A.~Tarvainen and H.~Valpola, ``Mean teachers are better role models: Weight-averaged consistency targets improve semi-supervised deep learning results,'' in \emph{Proc. Adv. Neural Inf. Process. Syst.}, 2017, pp. 1195--1204.

\bibitem{GrandvaletB04}
Y.~Grandvalet and Y.~Bengio, ``Semi-supervised learning by entropy minimization,'' in \emph{Proc. Adv. Neural Inf. Process. Syst.}, 2004, pp. 529--536.

\bibitem{lee2013pseudo}
D.-H. Lee \emph{et~al.}, ``Pseudo-label: The simple and efficient semi-supervised learning method for deep neural networks,'' in \emph{Proc. Int. Conf. Mach. Learn.}, vol.~3, no.~2, 2013, p. 896.

\bibitem{BerthelotCGPOR19}
D.~Berthelot, N.~Carlini, I.~J. Goodfellow, N.~Papernot, A.~Oliver, and C.~Raffel, ``Mixmatch: {A} holistic approach to semi-supervised learning,'' in \emph{Proc. Adv. Neural Inf. Process. Syst.}, 2019, pp. 5050--5060.

\bibitem{remixmatch}
D.~Berthelot, N.~Carlini, E.~D. Cubuk, A.~Kurakin, K.~Sohn, H.~Zhang, and C.~Raffel, ``Remixmatch: Semi-supervised learning with distribution alignment and augmentation anchoring,'' \emph{CoRR}, vol. abs/1911.09785, 2019.

\bibitem{SohnBCZZRCKL20}
K.~Sohn, D.~Berthelot, N.~Carlini, Z.~Zhang, H.~Zhang, C.~Raffel, E.~D. Cubuk, A.~Kurakin, and C.~Li, ``Fixmatch: Simplifying semi-supervised learning with consistency and confidence,'' in \emph{Proc. Adv. Neural Inf. Process. Syst.}, 2020, pp. 596--608.

\bibitem{XuSYQLSLJ21}
Y.~Xu, L.~Shang, J.~Ye, Q.~Qian, Y.~Li, B.~Sun, H.~Li, and R.~Jin, ``Dash: Semi-supervised learning with dynamic thresholding,'' in \emph{Proc. Int. Conf. Mach. Learn.}, vol. 139, 2021, pp. 11\,525--11\,536.

\bibitem{yang2021s2osc}
Y.~Yang, H.~Wei, Z.-Q. Sun, G.-Y. Li, Y.~Zhou, H.~Xiong, and J.~Yang, ``S2osc: A holistic semi-supervised approach for open set classification,'' \emph{ACM Trans. Knowl. Discov. Data}, vol.~16, no.~2, pp. 34:1--34:27, Sep. 2021.

\bibitem{zhao2022lassl}
Z.~Zhao, L.~Zhou, L.~Wang, Y.~Shi, and Y.~Gao, ``Lassl: label-guided self-training for semi-supervised learning,'' in \emph{Proc. AAAI Conf. Artif. Intell.}, vol.~36, no.~8, 2022, pp. 9208--9216.

\bibitem{saito2021openmatch}
K.~Saito, D.~Kim, and K.~Saenko, ``Openmatch: Open-set consistency regularization for semi-supervised learning with outliers,'' in \emph{Proc. Adv. Neural Inf. Process. Syst.}, 2021, pp. 25\,956--25\,967.

\bibitem{yang2024exploiting}
Y.~Yang, H.~Wei, H.~Zhu, D.~Yu, H.~Xiong, and J.~Yang, ``Exploiting cross-modal prediction and relation consistency for semisupervised image captioning,'' \emph{IEEE Trans. Cybern.}, vol.~54, no.~2, pp. 890--902, Feb. 2024.

\bibitem{yang2021semicluster}
Y.~Yang, D.-C. Zhan, Y.-F. Wu, Z.-B. Liu, H.~Xiong, and Y.~Jiang, ``Semi-supervised multi-modal clustering and classification with incomplete modalities,'' \emph{IEEE Trans. Knowl. Data Eng.}, vol.~33, no.~2, pp. 682--695, Feb. 2021.

\bibitem{guorobust}
L.-Z. Guo, Y.-G. Zhang, Z.-F. Wu, J.-J. Shao, and Y.-F. Li, ``Robust semi-supervised learning when not all classes have labels,'' in \emph{Proc. Adv. Neural Inf. Process. Syst.}, 2022, pp. 3305--3317.

\bibitem{xi2023robust}
W.~Xi, X.~Song, W.~Guo, and Y.~Yang, ``Robust semi-supervised learning for self-learning open-world classes,'' in \emph{Proc. IEEE Int. Conf. Data Min.}, 2023, pp. 658--667.

\bibitem{yang2023not}
Y.~Yang, Y.~Zhang, X.~SONG, and Y.~Xu, ``Not all out-of-distribution data are harmful to open-set active learning,'' in \emph{Proc. Adv. Neural Inf. Process. Syst.}, vol.~36, 2023, pp. 13\,802--13\,818.

\bibitem{ChenZLG20}
Y.~Chen, X.~Zhu, W.~Li, and S.~Gong, ``Semi-supervised learning under class distribution mismatch,'' in \emph{Proc. AAAI Conf. Artif. Intell.}, vol.~34, no.~04, 2020, pp. 3569--3576.

\bibitem{guo2020safe}
L.-Z. Guo, Z.-Y. Zhang, Y.~Jiang, Y.-F. Li, and Z.-H. Zhou, ``Safe deep semi-supervised learning for unseen-class unlabeled data,'' in \emph{Proc. Int. Conf. Mach. Learn.}, 2020, pp. 3897--3906.

\bibitem{yu2020multi}
Q.~Yu, D.~Ikami, G.~Irie, and K.~Aizawa, ``Multi-task curriculum framework for open-set semi-supervised learning,'' in \emph{Proc. Eur. Conf. Comput. Vis.}, 2020, pp. 438--454.

\bibitem{Wang0HHFW0SSRS023}
Y.~Wang, H.~Chen, Q.~Heng, W.~Hou, Y.~Fan, Z.~Wu, J.~Wang, M.~Savvides, T.~Shinozaki, B.~Raj, B.~Schiele, and X.~Xie, ``Freematch: Self-adaptive thresholding for semi-supervised learning,'' in \emph{Proc. Int. Conf. Learn. Representations}, 2023.

\bibitem{krizhevsky2009learning}
A.~Krizhevsky, G.~Hinton \emph{et~al.}, ``Learning multiple layers of features from tiny images,'' \emph{Tech. Rep.}, 2009.

\bibitem{yao2015tiny}
L.~Yao and J.~Miller, ``Tiny imagenet classification with convolutional neural networks,'' \emph{CS 231N}, vol.~2, no.~5, p.~8, 2015.

\bibitem{LeCunBH15}
Y.~LeCun, Y.~Bengio, and G.~E. Hinton, ``Deep learning,'' \emph{Nat.}, vol. 521, no. 7553, pp. 436--444, 2015.

\bibitem{XieDHL020}
Q.~Xie, Z.~Dai, E.~H. Hovy, T.~Luong, and Q.~Le, ``Unsupervised data augmentation for consistency training,'' in \emph{Proc. Adv. Neural Inf. Process. Syst.}, 2020, pp. 6256--6268.

\bibitem{ZhangWHWWOS21}
B.~Zhang, Y.~Wang, W.~Hou, H.~Wu, J.~Wang, M.~Okumura, and T.~Shinozaki, ``Flexmatch: Boosting semi-supervised learning with curriculum pseudo labeling,'' in \emph{Proc. Adv. Neural Inf. Process. Syst.}, 2021, pp. 18\,408--18\,419.

\bibitem{realmix2019}
V.~Nair, J.~F. Alonso, and T.~Beltramelli, ``Realmix: Towards realistic semi-supervised deep learning algorithms,'' \emph{CoRR}, vol. abs/1912.08766, 2019.

\bibitem{HuangF0CWWLL21}
J.~Huang, C.~Fang, W.~Chen, Z.~Chai, X.~Wei, P.~Wei, L.~Lin, and G.~Li, ``Trash to treasure: Harvesting {OOD} data with cross-modal matching for open-set semi-supervised learning,'' in \emph{{Proc. IEEE Int. Conf. Comput. Vis}}, 2021, pp. 8290--8299.

\bibitem{ParkYJS22}
J.~Park, S.~Yun, J.~Jeong, and J.~Shin, ``Opencos: Contrastive semi-supervised learning for handling open-set unlabeled data,'' in \emph{Proc. Eur. Conf. Comput. Vis.}, vol. 13802, 2022, pp. 134--149.

\bibitem{huang2022they}
Z.~Huang, J.~Yang, and C.~Gong, ``They are not completely useless: Towards recycling transferable unlabeled data for class-mismatched semi-supervised learning,'' \emph{{IEEE} Trans. Multim.}, vol.~25, pp. 1844--1857, Jun. 2023.

\bibitem{li2023iomatch}
Z.~Li, L.~Qi, Y.~Shi, and Y.~Gao, ``Iomatch: Simplifying open-set semi-supervised learning with joint inliers and outliers utilization,'' in \emph{Proc. IEEE Int. Conf. Comput. Vis}, 2023, pp. 15\,870--15\,879.

\bibitem{yangopenood}
J.~Yang, P.~Wang, D.~Zou, Z.~Zhou, K.~Ding, W.~Peng, H.~Wang, G.~Chen, B.~Li, Y.~Sun, X.~Du, K.~Zhou, W.~Zhang, D.~Hendrycks, Y.~Li, and Z.~Liu, ``Openood: Benchmarking generalized out-of-distribution detection,'' in \emph{Proc. Adv. Neural Inf. Process. Syst.}, vol.~35, 2022, pp. 32\,598--32\,611.

\bibitem{DuWCL22}
X.~Du, Z.~Wang, M.~Cai, and Y.~Li, ``{VOS:} learning what you don't know by virtual outlier synthesis,'' in \emph{{Proc. Int. Conf. Learn. Representations}}, 2022.

\bibitem{hendrycksbaseline}
D.~Hendrycks and K.~Gimpel, ``A baseline for detecting misclassified and out-of-distribution examples in neural networks,'' in \emph{{Proc. Int. Conf. Learn. Representations}}, 2017.

\bibitem{Shreyas2020ova}
S.~Padhy, Z.~Nado, J.~Ren, J.~Z. Liu, J.~Snoek, and B.~Lakshminarayanan, ``Revisiting one-vs-all classifiers for predictive uncertainty and out-of distribution detection in neural networks,'' \emph{CoRR}, vol. abs/2007.05134, 2020.

\bibitem{sun2022out}
Y.~Sun, Y.~Ming, X.~Zhu, and Y.~Li, ``Out-of-distribution detection with deep nearest neighbors,'' in \emph{{Proc. Int. Conf. Mach. Learn.}}, vol. 162, 2022, pp. 20\,827--20\,840.

\bibitem{WeiXCF0L22}
H.~Wei, R.~Xie, H.~Cheng, L.~Feng, B.~An, and Y.~Li, ``Mitigating neural network overconfidence with logit normalization,'' in \emph{Proc. Int. Conf. Mach. Learn.}, vol. 162, 2022, pp. 23\,631--23\,644.

\bibitem{yang2023learning}
Y.~Yang, Z.-Q. Sun, H.~Zhu, Y.~Fu, Y.~Zhou, H.~Xiong, and J.~Yang, ``Learning adaptive embedding considering incremental class,'' \emph{IEEE Trans. Knowl. Data Eng.}, vol.~35, no.~3, pp. 2736--2749, Mar. 2023.

\bibitem{HendrycksMD19}
D.~Hendrycks, M.~Mazeika, and T.~G. Dietterich, ``Deep anomaly detection with outlier exposure,'' in \emph{{Proc. Int. Conf. Learn. Representations}}, 2019.

\bibitem{YangWFYZZ021}
J.~Yang, H.~Wang, L.~Feng, X.~Yan, H.~Zheng, W.~Zhang, and Z.~Liu, ``Semantically coherent out-of-distribution detection,'' in \emph{Proc. IEEE Int. Conf. Comput. Vis}, 2021, pp. 8281--8289.

\bibitem{YuA19}
Q.~Yu and K.~Aizawa, ``Unsupervised out-of-distribution detection by maximum classifier discrepancy,'' in \emph{Proc. IEEE Int. Conf. Comput. Vis}, 2019, pp. 9517--9525.

\bibitem{cubuk2020randaugment}
E.~D. Cubuk, B.~Zoph, J.~Shlens, and Q.~V. Le, ``Randaugment: Practical automated data augmentation with a reduced search space,'' in \emph{{Proc. IEEE Conf. Comput. Vis. Pattern Recog.}}, 2020, pp. 3008--3017.

\bibitem{ghadimi2013stochastic}
S.~Ghadimi and G.~Lan, ``Stochastic first-and zeroth-order methods for nonconvex stochastic programming,'' \emph{SIAM J. Optim.}, vol.~23, no.~4, pp. 2341--2368, 2013.

\bibitem{bertsekas1995neuro}
D.~P. Bertsekas and J.~N. Tsitsiklis, ``Neuro-dynamic programming: an overview,'' in \emph{Proc. IEEE Conf. Decis. Control}, vol.~1, 1995, pp. 560--564.

\bibitem{bottou2018optimization}
L.~Bottou, F.~E. Curtis, and J.~Nocedal, ``Optimization methods for large-scale machine learning,'' \emph{SIAM Rev Soc Ind Appl Math}, vol.~60, no.~2, pp. 223--311, 2018.

\bibitem{opac-b1104789}
Y.~Nesterov, \emph{Introductory lectures on convex optimization : a basic course}, ser. Applied optimization.\hskip 1em plus 0.5em minus 0.4em\relax Kluwer Academic Publ., 2004.

\bibitem{polyak1963gradient}
B.~T. Polyak, ``Gradient methods for minimizing functionals,'' \emph{Zhurnal Vychislitel'noi Matematiki i Matematicheskoi Fiziki}, vol.~3, no.~4, pp. 643--653, 1963.

\bibitem{allen2019convergence}
Z.~Allen-Zhu, Y.~Li, and Z.~Song, ``A convergence theory for deep learning via over-parameterization,'' in \emph{Proc. Int. Conf. Mach. Learn.}, 2019, pp. 242--252.

\bibitem{Yuan0JY19}
Z.~Yuan, Y.~Yan, R.~Jin, and T.~Yang, ``Stagewise training accelerates convergence of testing error over {SGD},'' in \emph{Proc. Adv. Neural Inf. Process. Syst.}, 2019, pp. 2604--2614.

\bibitem{KarimiNS16}
H.~Karimi, J.~Nutini, and M.~Schmidt, ``Linear convergence of gradient and proximal-gradient methods under the polyak-{\l}ojasiewicz condition,'' in \emph{Lect. Notes Comput. Sci.}, vol. 9851, 2016, pp. 795--811.

\bibitem{abs-2002-05273}
X.~Li, Z.~Zhuang, and F.~Orabona, ``Exponential step sizes for non-convex optimization,'' \emph{CoRR}, vol. abs/2002.05273, 2020.

\bibitem{CharlesP18}
Z.~Charles and D.~S. Papailiopoulos, ``Stability and generalization of learning algorithms that converge to global optima,'' in \emph{Proc. Int. Conf. Mach. Learn.}, vol.~80, 2018, pp. 744--753.

\bibitem{LiL18}
Z.~Li and J.~Li, ``A simple proximal stochastic gradient method for nonsmooth nonconvex optimization,'' in \emph{Proc. Adv. Neural Inf. Process. Syst.}, 2018, pp. 5569--5579.

\bibitem{otsu1979threshold}
N.~Otsu, ``A threshold selection method from gray-level histograms,'' \emph{IEEE Trans. Syst. Man Cybern.}, vol.~9, no.~1, pp. 62--66, Jan. 1979.

\bibitem{CaoBL22}
K.~Cao, M.~Brbic, and J.~Leskovec, ``Open-world semi-supervised learning,'' in \emph{{Proc. Int. Conf. Learn. Representations}}, 2022.

\bibitem{onpseudo2023}
L.~Han, H.~Ye, and D.~Zhan, ``On pseudo-labeling for class-mismatch semi-supervised learning,'' \emph{CoRR}, vol. abs/2301.06010, 2023.

\bibitem{DengDSLL009}
J.~Deng, W.~Dong, R.~Socher, L.~Li, K.~Li, and L.~Fei{-}Fei, ``Imagenet: {A} large-scale hierarchical image database,'' in \emph{Proc. IEEE Conf. Comput. Vis. Pattern Recog.}, 2009, pp. 248--255.

\bibitem{DavisG06}
J.~Davis and M.~Goadrich, ``The relationship between precision-recall and {ROC} curves,'' in \emph{{Proc. Int. Conf. Mach. Learn.}}, W.~W. Cohen and A.~W. Moore, Eds., vol. 148, 2006, pp. 233--240.

\bibitem{ZagoruykoK16}
S.~Zagoruyko and N.~Komodakis, ``Wide residual networks,'' in \emph{Proc. Br. Mach. Vis. Conf.}, 2016.

\bibitem{tack2020csi}
J.~Tack, S.~Mo, J.~Jeong, and J.~Shin, ``{CSI:} novelty detection via contrastive learning on distributionally shifted instances,'' in \emph{Proc. Adv. Neural Inf. Process. Syst.}, vol.~33, 2020, pp. 11\,839--11\,852.

\bibitem{YuZSSX15}
F.~Yu, Y.~Zhang, S.~Song, A.~Seff, and J.~Xiao, ``{LSUN:} construction of a large-scale image dataset using deep learning with humans in the loop,'' \emph{CoRR}, vol. abs/1506.03365, 2015.

\bibitem{CimpoiMKMV14}
M.~Cimpoi, S.~Maji, I.~Kokkinos, S.~Mohamed, and A.~Vedaldi, ``Describing textures in the wild,'' in \emph{Proc. IEEE Conf. Comput. Vis. Pattern Recog.}, 2014, pp. 3606--3613.

\bibitem{wah2011caltech}
C.~Wah, S.~Branson, P.~Welinder, P.~Perona, and S.~Belongie, ``The caltech-ucsd birds-200-2011 dataset,'' \emph{Tech. Rep.}, 2011.

\bibitem{NilsbackZ06}
M.~Nilsback and A.~Zisserman, ``A visual vocabulary for flower classification,'' in \emph{{Proc. IEEE Conf. Comput. Vis. Pattern Recog.}}, vol.~2, 2006, pp. 1447--1454.

\bibitem{griffin2007caltech}
G.~Griffin, A.~Holub, and P.~Perona, ``Caltech-256 object category dataset,'' \emph{Tech. Rep.}, 2007.

\bibitem{khosla2011novel}
A.~Khosla, N.~Jayadevaprakash, B.~Yao, and F.-F. Li, ``Novel dataset for fine-grained image categorization: Stanford dogs,'' in \emph{Proc. IEEE Conf. Comput. Vis. Pattern Recog.}, vol.~2, no.~1, 2011.

\end{thebibliography}

\begin{IEEEbiography}[{\includegraphics[width=1in,height=1.25in,clip,keepaspectratio]{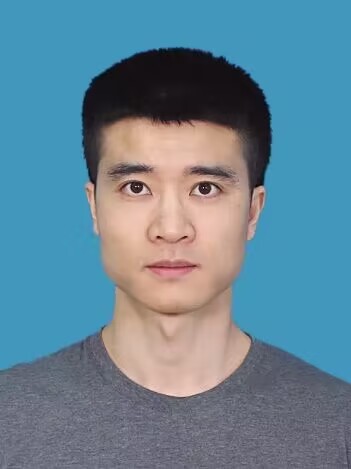}}]{Yang Yang}
    received the Ph.D. degree in computer science, Nanjing University, China in 2019. At the same year, he became a faculty member at Nanjing University of Science and Technology, China. He is currently a Professor with the school of Computer Science and Engineering. His research interests lie primarily in machine learning and data mining, including heterogeneous learning, model reuse, and incremental mining.  He has published prolifically in refereed journals and conference proceedings, including IEEE Transactions on Knowledge and Data Engineering (TKDE), ACM Transactions on Information Systems (ACM TOIS), ACM Transactions on Knowledge Discovery from Data (TKDD), ACM SIGKDD, ACM SIGIR, WWW, IJCAI, and AAAI. He was the recipient of the the Best Paper Award of ACML-2017. He serves as PC in leading conferences such as IJCAI, AAAI, ICML, NeurIPS, etc.
\end{IEEEbiography}
\begin{IEEEbiography}[{\includegraphics[width=1in,height=1.25in,clip,keepaspectratio]{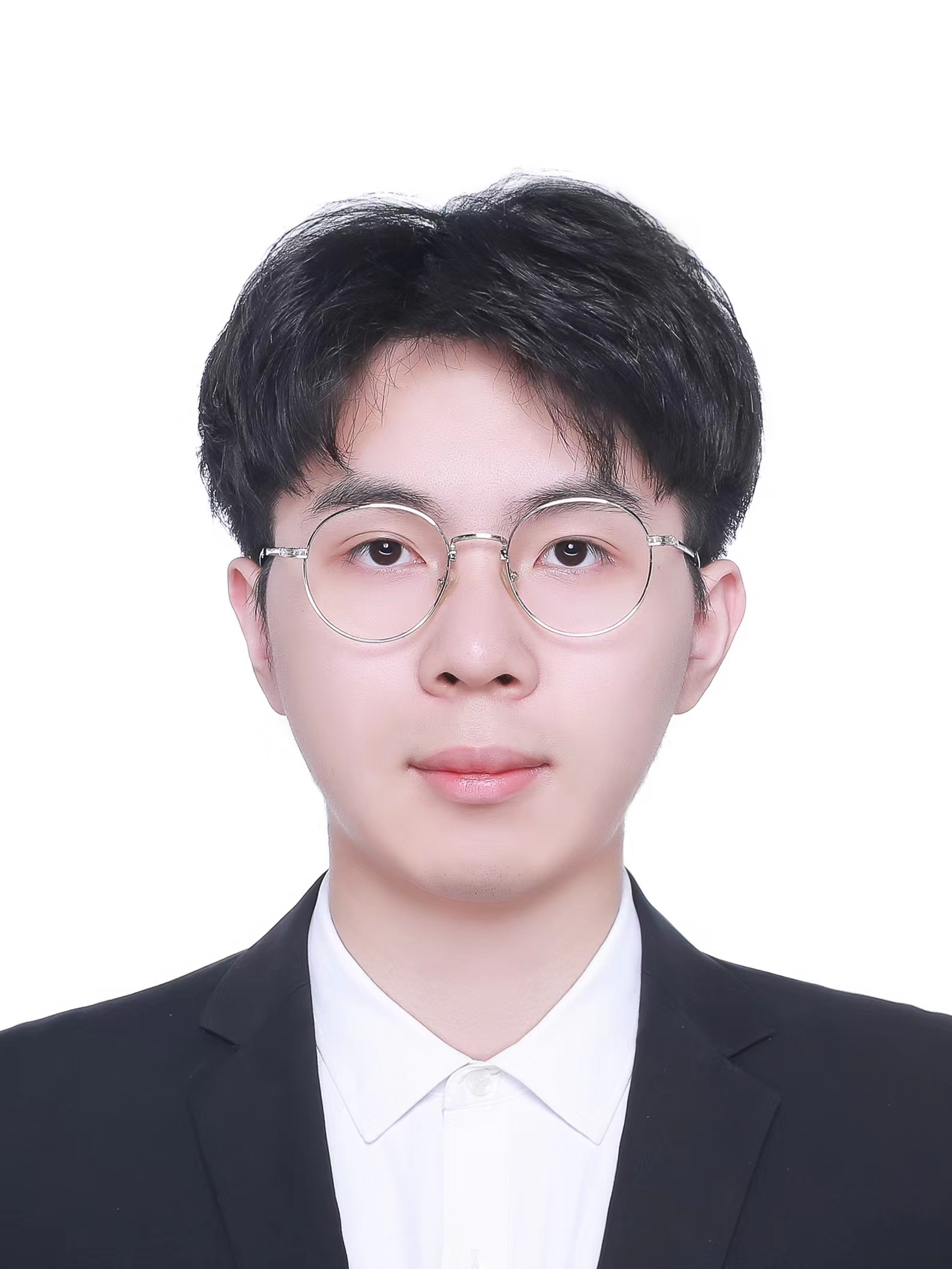}}]{Nan Jiang}
    is currently working towards the M.S. degree with the National Key Laboratory for Novel Software Technology, School of Artificial Intelligence, Nanjing University, China. His research interests lie primarily in machine learning and data mining. He is currently working on semi-supervised learning.
\end{IEEEbiography}
\begin{IEEEbiography}[{\includegraphics[width=1in,height=1.25in,clip,keepaspectratio]{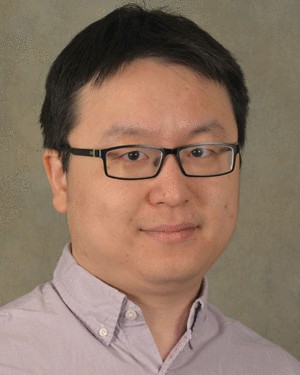}}]{Yi Xu}
    received the PhD degree in computer science from the University of Iowa, Iowa City, Iowa USA, in 2019. He is currently a professor with the School of Control Science and Engineering, Dalian University of Technology, China. His research interests are machine learning, optimization, deep learning, and statistical learning theory. He has published more than twenty papers in refereed journals and conference proceedings, including IEEE Transactions on Pattern Analysis and Machine Intelligence (TPAMI), Transactions on Machine Learning Research, NeurIPS, ICML, ICLR, CVPR, AAAI, IJCAI, UAI. He serves as a SPC/PC/Reviewer in leading conferences such as NeurIPS, ICML, ICLR, CVPR, AAAI, IJCAI, etc.     
\end{IEEEbiography}
\begin{IEEEbiography}[{\includegraphics[width=1in,height=1.25in,clip,keepaspectratio]{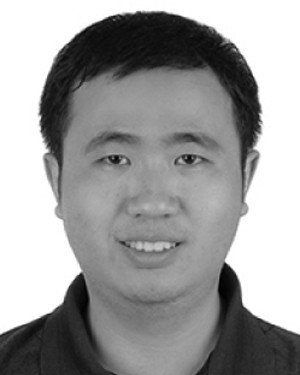}}]{De-Chuan Zhan}
    received the PhD degree in computer science, Nanjing University, China in 2010, then he became a faculty member with the Department of Computer Science and Technology at Nanjing University, China. He is currently a professor with the School of Artificial Intelligence at Nanjing University. His research interests are mainly in machine learning, data mining, and mobile intelligence. He has published more than 90 papers in leading international journal/conferences. He serves as an editorial board member of IDA and IJAPR, and serves as SPC/PC in leading conferences, such as IJCAI, AAAI, ICML, NeurIPS, etc.
\end{IEEEbiography}


\ifCLASSOPTIONcaptionsoff
  \newpage
\fi

\end{document}